\documentclass[preprint,review, 12pt]{elsarticle}

\usepackage{xcolor}
\usepackage{nicefrac}
\usepackage{amssymb}
\usepackage{lineno} 
\usepackage[mathscr]{eucal} 
\usepackage{amsmath} 
\usepackage{algorithmic} 
\usepackage[linesnumbered,lined,boxed,commentsnumbered,ruled]{algorithm2e}
\usepackage{booktabs}
\usepackage{multirow} 
\usepackage{lscape} 
\usepackage{setspace}
\RequirePackage[pdftex,
      colorlinks={true},
      linkcolor={Blue3},
      urlcolor={Blue3},
      citecolor={Blue3},
      pdfstartview={FitH},
      bookmarks={true},
            pdfauthor={Author},
            pdftitle={Title},
            pdfsubject={Subject}
            plainpages=false,
            pdfpagelabels
            ]{hyperref}

\journal{Neurocomputing}

\begin{document}

\begin{frontmatter}

\title{nTreeClus: a Tree-based Sequence Encoder for Clustering Categorical Series}

\author[1]{Hadi Jahanshahi$^{*,}$} 

\ead{hadi.jahanshahi@ryerson.ca}

\address[1]{Data Science Lab at Ryerson University, Toronto, ON M5B 1G3, Canada}

\author[2]{Mustafa Gokce Baydogan} 
\ead{mustafa.baydogan@boun.edu.tr}
\ead[URL]{www.mustafabaydogan.com}

\address[2]{Department of Industrial Engineering: Boğaziçi University, 34342, Bebek, Istanbul, Turkey}

\cortext[cor1]{Corresponding author}

\begin{abstract}
The overwhelming presence of categorical/sequential data in diverse domains emphasizes the importance of sequence mining. The challenging nature of sequences proves the need for continuing research to find a more accurate and faster approach providing a better understanding of their (dis)similarities. This paper proposes a new Model-based approach for clustering sequence data, namely nTreeClus. The proposed method deploys Tree-based Learners, $k$-mers, and autoregressive models for categorical time series, culminating with a novel numerical representation of the categorical sequences. Adopting this new representation, we cluster sequences, considering the inherent patterns in categorical time series. Accordingly, the model showed robustness to its parameter. Under different simulated scenarios, nTreeClus improved the baseline methods for various internal and external cluster validation metrics for up to 10.7\% and 2.7\%, respectively. The empirical evaluation using synthetic and real datasets, protein sequences, and categorical time series showed that nTreeClus is competitive or superior to most state-of-the-art algorithms. 
\end{abstract}

\begin{highlights}
\item A novel model-based clustering approach for sequential data is proposed.

\item The proposed method introduces Decision Tree Path encoder implemented in an autoregressive manner.

\item The method's robustness to its only parameter (window size) has been examined.

\item The approach is tested on multiple datasets and provides competitive results compared to existing methods in sequence mining.
\end{highlights}


\end{frontmatter}
\newpage
\section{Introduction}
\label{sec:intro}

Sequences exist in diverse domains such as bioinformatics, marketing, social science, text mining, web mining, security, health-care, or business. For example, short repeated DNA sequences are one of the most intriguing features of prokaryotic and eukaryotic genomes since they decipher the structure and mechanism associated with neurological disorders~\cite{Karaca2005}. 
Moreover, click-stream sequences that capture customers' preferences through their online purchase history are another predominant type of categorical sequences~\cite{
Montgomery2004}. Analyzing the past transaction logs or historical data of an organization is of importance for decision-makers~\cite{masseglia2005sequential}. Assessing process flows in a hospital (as a sequence of past events)~
and evaluating the past transaction data of a company's orders ~
are some other typical examples of the ubiquity of sequences. 

Although there have been numerous studies on sequence data mining, clustering categorical sequences still needs more attention in the literature noting its usefulness for a plethora of domains. The task demands either accurately computing (dis)similarity between sequences or finding the underlying model generating sequences. Therefore, from the problem-solving perspective, we categorize them into two folds: \textit{Proximity-Based approaches}, including \textit{Similarity-based} and \textit{Feature-based}; and \textit{Model-based approaches}~\cite{similaritybased}. The main aim of Proximity-Based methods is to devise a similarity or distance measure for sequences, applied either on raw-data in \textit{Similarity-based} or on feature-extracted data in \textit{Feature-based}. On the other hand, in \textit{Model-based} methods, a new representation of the given data has been introduced through a model that best fits the data. Hence, to cluster categorical time series, we need to define a (dis)similarity measure or latent model that appropriately depicts the behavior of sequences. 

There has been a significant change in approaches to solving these types of problems during the past decade. Previously, Similarity-based analyses, especially alignment-based ones, were the most popular methods, aligning sequences and finding the best or longest common pattern that exists in a sequence. They can be categorized as global, local, or glocal alignments~\cite{Glocalalignment1}, a pairwise alignment used for only two sequences at a time~\cite{BlastAltschul97}, multi-sequence alignment~\cite{EDGAR2006368}, and structural alignment that applies to sequences whose structures are known~\cite{holm2010dali}. Accordingly, most of these methods are used to handle bioinformatics sequences and, in most cases, cannot be generalized to other domains. Additionally, most of the more accurate ones have many limiting factors, namely, the limitation on the number of sequences, computation time and memory, pattern position in a sequence, and the sequence length that should not be varying significantly.

Similarity-based approaches considering distance among sequences are apt to disclose the (dis)similarity between given sequences. Most of the methods in this category use the cost of dissimilarity to count mismatches; nevertheless, the definition of mismatch in each algorithm differs. Hamming distance, applicable to the sequence with the same length, counts the number of elements substituted to change one string to another. The Levenshtein distance, or edit distance, is the number of insertions, deletions, and replacements required to turn one string into another~\cite{burkhardt2002one}. The method is not a proper choice due to its computational complexity, its median string is not tractable~\cite{dinu2006low}, and, finally, it fails in some significant cases such as automated language classification tasks~\cite{Greenhill2011}.       

Feature-based approaches perform information abstraction by capturing merely important information about the data. $k$-mers ($n$-grams), as the most common method in this category, are more omnipresent than alignment methods. They are adopted in word prediction (speech recognition), text categorization and classification, similarity and distance of strings, authorship identification, protein and genome sequences, and information retrieval. Nevertheless, there are some arguments about its robustness to its parameter, i.e., finding the optimal $k$ for $k$-mer~\cite{paynabar2016sequence}. Position Weight Matrix (PWM) is another method that finds a pattern, especially a motif, in a sequence~\cite{stormo1982use}. It uses a matrix representation of a sequence based on the appearance of each element in a specific window. This method requires pre-alignment for the sequences with different lengths to find the optimal window. Hence, the problem of finding the best window size arises.


In Model-based approaches, a model is trained to understand the underlying mechanism that generates the categorical sequence. Markovian models assume that sequences have parametric distributions. They heavily depend on the conditional probability of occurrence of each element in a sequence while satisfying the so-called ``Markov Property''. There are many types of Markovian Models, including first-order Markov Chain, Higher-order Markov Chain, Interpolated Markov Motif Models (IMM), Multivariate Markov Chain Model, and Hidden Markov Model (HMM) and its Mixture version (MHMM). Model-based methods, such as MHMM, simultaneously cluster sequences while finding the best-fitted model. Unlike PWM that has no memory, Markov Chain Models consider previous elements of the sequence by introducing states~\cite{dong2007sequence}. Despite being utilized in diverse domains, HMMs are hard to be tuned and, in most cases, are computationally intensive. This fact affects its generalizability and scalability~\cite{paynabar2016sequence}. On the other hand, AutoRegressive models (AR models) investigate the correlative structure of sequences. However, studies on AR Models for sequential data are mostly applicable to bioinformatics and lack generalizability. They also require setting model order and window length. Finding the optimal parameters takes time and may lead to the higher complexity of the model~\cite{Akhtar4449750}. AR models also require numerical mapping, while categorical data might not have a logical order, and mapping them may lead to an irrational deduction.

In this study, we propound a new model-based clustering method to capture the sequence features while overcoming its counterparts' barriers. For brevity, we call this framework as nTreeClus in the following context. nTreeClus offers a novel encoding of categorical sequential data fed to \linebreak(dis)similarity measures for clustering sequences. it combines the notions behind $k$-mers, AR models, tree-based algorithms, and similarity measures to generate a new comprehensive method. Accordingly, it segments a given sequence into shorter subsequences of the length $n$. Afterward, it traces back the relation among all elements of the sequence through an autoregressive Decision Tree (DT). We count the number of repetitions of each segmented subsequence in terminal nodes (leaf nodes) of DTs to create a new numeric representation of the sequence. Hence, available (dis)similarity methods, e.g., Cosine or Manhattan distance, can be used to cluster them. The experiments with synthetic and real data reveal the robustness and accuracy of the method.

The remainder of the paper is organized as follows. Section~\ref{sec:background} summarizes a brief notion of the method, \textcolor{black}{and its competitive approaches are explained in Section~\ref{sec:related_works}.} In Section~\ref{sec:method}, the preliminaries are introduced, and nTreeClus as a method to cluster sequential data is presented, including the novel approach for representing categorical data and similarity measurements. Furthermore, the computational complexity of the method is briefly discussed. Section~\ref{sec:experiments} discusses the sensitivity of the method to its parameter $n$ and its ability to predict the number of clusters. Moreover, it reports the performance of the approach under different scenarios and examines its clustering ability on the real data. \textcolor{black}{Section~\ref{sec:discussion} explains the threats to validity and possible future extensions, and Section~\ref{sec:conclusion} conclude the paper.}

\section {Methodology}
\subsection {Background}
\label{sec:background}
nTreeClus synthesizes the ideas of tree-based learners, $k$-mers, and AR models for categorical time series. This section elucidates the concept behind these algorithms and models, in addition to their virtues and shortcomings. 

Decision Tree learners are interpretable models with satisfactory accuracy used in a wide variety of domains. Univariate Decision Trees, such as CART~\cite{breiman1984CART} and C4.5~\cite{quinlan1993c4}, regard only one attribute (feature) at each decision node, leading to axis-aligned splits. Unlike their greedy nature, tree ensembles, e.g., Random Forest (RF), are less likely to overfit on data --- they are a collection of DTs whose results are less biased and less likely to overfit due to the random feature selection. RF classifier is an ensemble of $k$ decision trees, $h(x,\Theta_k)$, where the $\Theta_k$ is a random vector which is independent of past vectors but with the same distribution, each casting a vote for the most popular class at input $x$~\cite{Breiman2001}. Often $\sqrt{\nu}$ is considered as the number of features to be used, where $\nu$ is the number of total available features. The random selection of features leads to the reduction of variance of the classifier and decreases computational complexity of DT from $O(\nu \mathscr{N} \log{\mathscr{N}})$ to $O(\sqrt{\nu} \mathscr{N}\log{\mathscr{N}})$ where $\mathscr{N}$ is the number of training instances (in-bag). RFs are simple to interpret and implement, robust to outliers, strong in handling non-linearity, fast in real cases, and friendly in parallel training to save more time~\cite{refinementofRF, Biau2016}. 

AR models are widely employed in data mining domains, especially time series, indicating the output variable depends on its previous values. Due to their stochastic nature, they take advantage of the statistics of the data~\cite{ARmodel}. Not until recently have data scientists been engaged in applying AR models to categorical sequences. AR spectral analysis tools capture certain features of DNA sequences of coding and non-coding regions~\cite{Chakravarthy2004AMF}. It claims that, based on the type of numerical mapping rule, the AR Feature-based searching techniques have a high accuracy in identifying the patterns and their locations in DNA sequences. Later on, many scholars verify that AR models for categorical time-series perform well in analyzing the structure of sequences and finding patterns in them~\cite{Akhtar4449750, Zhou4274026, Rosen4365814, BLINOWSKA2009104, Choong4811253, Song1946489}. However, AR models are new and mostly implemented in bioinformatics, thus lacking generalizability. They also have two hyperparameters, i.e., model order and window length, leading to their higher complexity. Finally, they use a numerical mapping of categorical data that leads to an irrational deduction.

\subsection{Related works}\label{sec:related_works}
In this section, we briefly introduce the popular methods from each of the three available approaches, including Jaro-Winkler and Levenshtein from Similarity-based methods, $k$-mers from Feature-based approaches, and finally, a Mixture of Hidden Markov Models from Model-based approaches. They are our baselines for pattern recognition in sequences. 

Jaro metric~\cite{jaro1989}, widely used in the record-linkage community, has a variation due to Winkler~\cite{Winkler99thestate}, counting the common characters between strings even if their positions are slightly different. This popular edit-distance method emphasizes the strings that match from the beginning (common prefix); therefore, it is unstable from one set to another~\cite{Giorgos}.  Levenshtein distance (LD)~\cite{levenshtein1966binary}, edit distance, is a precise measurement to compare two or more strings by various operations, including insertions, deletions, and reversals (substitutions) of symbols. It relies on the minimum cost to transform one string into another through necessary modifications. The more edit operations we use for transformation, the more distant two strings are. Its computational complexity increases as the size of two strings increases. 

$n$-gram method, also called $k$-mer in bioinformatics, is a Feature-based method used as a baseline since the logic behind window size ($n$ in $n$-gram and $k$ in $k$-mer) is closely comparable to $n$ in nTreeClus. Mixture Hidden Markov models (MHMM), also called mixed Markov latent class models~\cite{MixedMarkov} or mixture latent Markov Model (MLMM)~\cite{vermunt2008latent}, contains five types of variables: response variables, time-constant explanatory variables, time-varying explanatory variables, time-constant discrete latent variables, and time-varying discrete latent variables. Here, for the simplicity of the model, we only consider time-constant covariates. Having a set of Hidden Markov Models $\mathscr{L} = \{\mathscr{L}^1, ..., \mathscr{L}^K\}$, where $\mathscr{L}^i = \{\pi^i, A^i, B^i_1, ..., B^i_C\}$ for submodels $i = 1,..., I$. For each subject $Y_i$, $P(\mathscr{L}^i) = \omega_i$ is the prior probability that it follows the submodel $\mathscr{L}^i$. The log-likelihood of the parameters of the Mixture Hidden Markov Model can be formulated as~\cite{helske2017mixture}

\begin{equation}
\begin{array}{l}
\log{L} = \sum\limits_{i=1}^{N}\log{P(Y_i|\mathscr{L})} \\
\quad \quad = \sum\limits_{i=1}^{N}\log{\Big[\sum\limits_{k=1}^{N}{P\big(\mathscr{L}^k\big)\sum\limits_{all z} {P\big(Y_i|z,\mathscr{L}^k\big)P\big(z|\mathscr{L}^k\big)}}  \Big].}
\end{array}
\end{equation}

It assumes there exists a latent state resulting in the observed sequence. In all the experiments, we select the number of latent states ($m$) based on Bayesian information criterion (BIC), $BIC = -2l(\theta) + p \log{m}$ for $m \in [1,5]$. Minimizing the BIC that corresponds to maximizing the posterior model probability is desirable~\cite{BICmodelselection}. 

\paragraph{Contributions} Our proposed algorithm, nTreeClus, differs from the existing literature in that it learns a vectorial representation of sequences through an auto-regressive tree-based algorithm. \citet{Zhang2018} introduced the Tree2Vector algorithm by emphasizing the importance of tree structures. Tree2Vector is a learning framework that transforms any tree-structured data into vector space. It encodes a tree using a bottom-up procedure in which \mbox{$k$-means} clustering allocates nodes to different clusters for different levels of the tree. Therefore, they achieve a vectorial representation of the tree that can be used in various applications, e.g., book author recommendations. Unlike the similarity in methods' names, nTreeClus differs from Tree2Vector since our algorithm only uses the binary decision tree structure or its ensemble versions and aims to decode categorical sequences through their autoregressive relation. To the best of our knowledge, it is the first time that terminal node representation of decision trees is proposed for sequence decoding. 

\subsection{nTreeClus: A new methodology for clustering categorical sequential data}
\label{sec:method}
The categorical sequential dataset, $X^\mathscr{L}_\mathscr{N}$, is comprised of $\mathscr{N}$ sequences of length $\mathscr{L}$ over an alphabet set of $\mathscr{A}$ where $x^m_l$ is the $m$th element of sequence $l$. For simplification, the sequences are assumed to be of the same length, $\mathscr{L}$, while the approach can handle sequences of different length. When the length of sequences is different, $\mathscr{L}$ is equal to $max \{\mathscr{L}_1,\mathscr{L}_2,..,\mathscr{L}_l \}$ and shorter sequences will be padded by missing tokens. Therefore, $x^j_i$ can either be selected  of the alphabet set $\mathscr{A} = \{s_1,s_2,..,s_a\}$ or remain empty in the case that $\mathscr{L}_i < j$. $X_i$, the $i$th sequence of the dataset, can be represented by
\begin{equation}
X_i = x^1_i x^2_i .. x^{\mathscr{L}-1}_i x^\mathscr{L}_i
\end{equation}
where it forms the $i$th row of the categorical dataset, $X^\mathscr{L}_\mathscr{N}$; thus, $X^\mathscr{L}_\mathscr{N}$ is illustrated by an $\mathscr{N} \times{\mathscr{L}}$ matrix as

\begin{equation}
X^\mathscr{L}_\mathscr{N} = 
\begin{bmatrix}
x^1_1  & x^2_1  &  \cdots  & \cdots & x^{\mathscr{L}-1}_1  &  x^\mathscr{L}_1 \\
x^1_2  & x^2_2  &  \cdots  & \cdots & x^{\mathscr{L}-1}_2  &  x^\mathscr{L}_2 \\
\vdots & \vdots & \ddots &        &  \vdots &  \vdots \\
\vdots & \vdots &        & \ddots &  \vdots &  \vdots \\
x^1_{\mathscr{N}-1}  & x^2_{\mathscr{N}-1}  &  \cdots  & \cdots & x^{\mathscr{L}-1}_{\mathscr{N}-1} &  x^\mathscr{L}_{\mathscr{N}-1} \\
x^1_\mathscr{N}  & x^2_\mathscr{N}  &  \cdots  & \cdots & x^{\mathscr{L}-1}_\mathscr{N}  &  x^\mathscr{L}_\mathscr{N} \\
\end{bmatrix}
\label{equation:dataset}
\end{equation}

nTreeClus encodes categorical sequences to numeric values and then finds specific patterns in them. It is extensible to any sequence in which we are looking for a pattern. Accordingly, it utilizes basic principles in available and commonly used approaches. A pattern recognizer can cluster sequences in which the definite number of alphabets is used and can specify their distinctive characteristic. 

\paragraph{Phase I: Segmented matrix representation}
In the first step, nTreeClus segments each sequence using a predefined length ($n$) to have a length-smoothed dataset. This step provides an environment in which employing single tree-based algorithms is feasible. It also has the same logic as $k$-mers with the window size of $n$ while extracting all possible substrings of length $n$ in a given string. Based on the predefined length $n$, we divide each into substrings with the length of $n+1$ (Figure~\ref{fig:seg_seq}). The last column is used as a class label, whereas the first $n$ elements play the role of features resulting in the class label (in an autoregressive manner). In the given sequence in Figure~\ref{fig:seg_seq}, $n$ is equal to 5. ``a b c a a'' are the five elements leading to the output ``a'' that is the next alphabet of the string. The number of new substrings is determined by both the length of each string and the predetermined ``$n$''. In Algorithm~\ref{alg:algorithm1}, we demonstrate the steps to create the segmented matrix given the sequential dataset.  
\begin{figure} [!ht]
\centering
\begin{equation*}
X_1:abcaaabcbaa \xrightarrow[\text{}]{\text{windows size = $5$}}
\begin{array}{cccccc|c}
  1^* & a & b & c & a & a & a \\
  2^* & b & c & a & a & a & b  \\
  3^* & c & a & a & a & b & c \\
  4^* & a & a & a & b & c & b \\
  5^* & a & a & b & c & b & a \\
  6^* & a & b & c & b & a & a \\
\end{array}
\end{equation*}
\caption{An example of segmented sequence (the right column indicates labels, while the first six columns are features). Since the label and features come from the same data, it mimics autoregressive behavior. The far-most left column is the location of each substring. We include this optional column whenever the position of the pattern is of importance.}
\label{fig:seg_seq}
\end{figure}

We apply the segmentation phase to all rows of matrix $X^\mathscr{L}_\mathscr{N}$ and generate the segmented matrix. Figure~\ref{fig:seg_seq} demonstrates the segmentation algorithm only for $X_1$, the first row of the supposed $X^\mathscr{L}_\mathscr{N}$. Assuming all rows of $X^\mathscr{L}_\mathscr{N}$ have the same length of $\mathscr{L} = 11$ and the total number of sequences is $\mathscr{N} = 100$, then for the window size of $n = 5$, we will have the segmented matrix (S) with 600 rows ($(11-5)\times{100}$) and six columns. Generally speaking, the segmented output is a matrix of size $((\mathscr{L}-n) \times{\mathscr{N}})\text{ by }{(n+1)}$.

\begin{algorithm}[!ht]

	\SetKwData{SegMat}{Segmented Matrix}\SetKwData{Temp}{Temporary Row}\SetKwData{D}{Dataset}	\KwData{sequential/categorical dataset ($D$) of size $\mathscr{N} \times \mathscr{L}$, the segmentation length ($n$) (\textit{with the default value of $\sqrt{\mathscr{L}}$}).}
	\KwResult{Segmented matrix (length-smoothed sequences)}
    initialization;\\
	\emph{\SegMat = $\emptyset$}\
	\BlankLine
	\For{$i \in{\{1,\ldots,\mathscr{N}\}}$}{
		\For{$j\in{\{1,\ldots,(\mathscr{L}_i-n)\}}$}{
			\Temp$\leftarrow$ \D\![i,j:j+n]\
			
			\Temp\![`y']$\leftarrow i$\ 
			
			\begin{algorithmic}
    			\IF{the position of element is important}
    			\STATE \Temp\![`position']$\leftarrow$ $j$\
    			\ENDIF
			\end{algorithmic} 
			Append [\Temp\!]\ to \SegMat	
		}
	}
	\caption{Matrix Segmentation}
	\label{alg:algorithm1}
\end{algorithm}

\begin{figure} [!ht]
	\centering
	\includegraphics[width=0.7\linewidth]{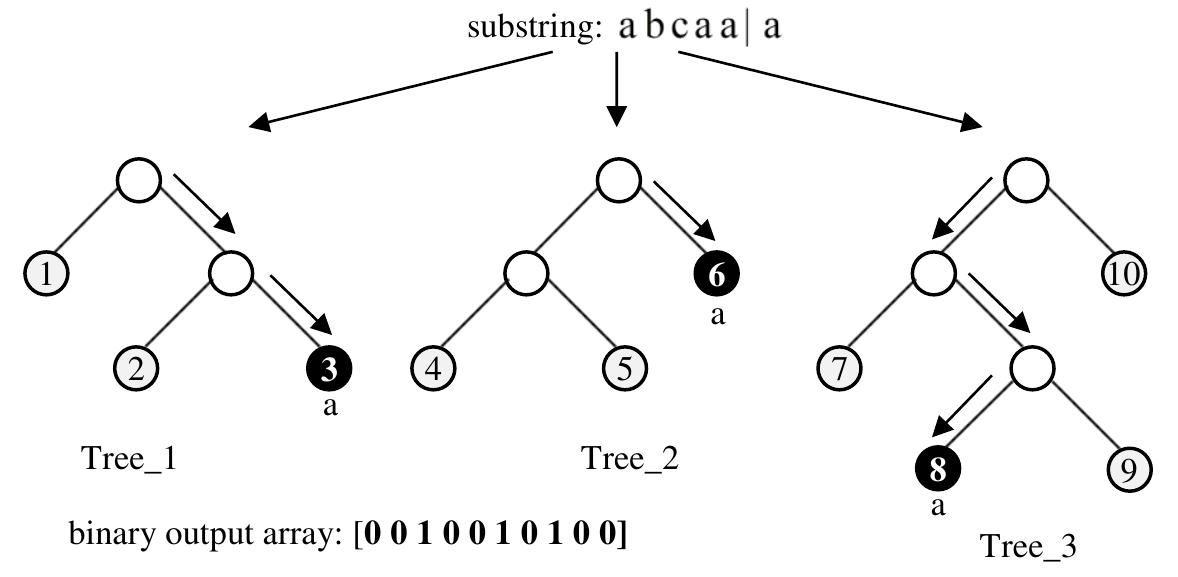}
	\caption{Introducing an instance to Decision Tree Ensembles}
	\label{fig:Rand_For}
\end{figure}

\paragraph{Phase II: Providing the novel nTreeClus representation}
The segmented matrix is an input for the next phase of the algorithm, where we use a DT. The CART DT, utilized by nTreeClus, is a binary partitioning process that can handle both continuous and discrete attributes. The Gini index supervises the splitting part so that the reduction in tree impurity becomes maximized~\cite{TppTen}. nTreeClus utilizes either DT or tree ensembles to minimize the error. In this stage, we encode each string via a set of terminal nodes' (leaf nodes') indices to which a specific substring. In other words, after implementing tree ensembles, each substring goes to a specific terminal node of each tree, and nTreeClus traces the exact position of the substring -– i.e., each rule of DT describes a pattern in the sequence. Assuming that the total number of trees in decision tree ensembles is three, the first substring in Figure~\ref{fig:seg_seq}, ``abcaa'', is assigned to terminal nodes 5, 3, and 6 of tree 1, 2, and 3, respectively (Figure~\ref{fig:Rand_For}). The other substrings experience the same flow.

nTreeClus records a large, sparse matrix of terminal nodes to which each instance belongs. Therefore, the number of its columns is equal to the number of terminal nodes, here 10, and the number of its rows is equal to the number of strings (instances) given by the original matrix. For the given sample in Figure~\ref{fig:seg_seq}, six substrings end in the terminal nodes, as shown in Equation~\ref{equation:sparsematrix}. For instance, the first row of this binary matrix shows that the substring $abcaa$ ends in terminal 2, 6, and 8 of the tree 1, 2, and 3, respectively.


\begin{equation}
\eta_{(abcaaabcbaa)} = 
\begin{pmatrix} 
	0 & 0 & 1 & 0 & 0 & 1 & 0 & 1 & 0 & 0 \\
	0 & 1 & 0 & 0 & 1 & 0 & 0 & 0 & 0 & 1 \\
	1 & 0 & 0 & 0 & 1 & 0 & 0 & 0 & 0 & 1 \\
	1 & 0 & 0 & 1 & 0 & 0 & 1 & 0 & 0 & 0 \\
	0 & 0 & 1 & 1 & 0 & 0 & 0 & 1 & 0 & 0 \\
	0 & 1 & 0 & 0 & 0 & 1 & 0 & 0 & 0 & 1 
\end{pmatrix}
\label{equation:sparsematrix}
\end{equation}

Equation~\ref{equation:sparsematrix} holds the terminal IDs for the sequence ``abcaaabcbaa", defined by its six substrings. It can be aggregated as the Equation~\ref{equation:aggregated}, whose number of columns is equal to the number of terminal nodes. This aggregated form demonstrates the frequency that each instance ends in a specific terminal node of DT Ensembles. ``0'' under column nine shows none of the substrings of the sequence ``abcaaabcbaa" is ended in terminal 9. Nonetheless, the number of rows of Equation~\ref{equation:aggregated} will be equal to the total number of instances, one of which is illustrated herein.
\begin{equation}
\eta_{aggregated} = 
\begin{pmatrix} 
2 & 2 & 2 & 2 & 2 & 2 & 1 & 2 & 0 & 3. \\
\end{pmatrix}
\label{equation:aggregated}
\end{equation}

Algorithm~\ref{alg:nTreeClusRepresentation} shows the encoding procedure for a given sequence. In this example, nTreeClus encodes the categorical sequence (abcaaabcbaa) into (2222221203), which maintains the meaning behind this set of alphabets.

\begin{algorithm}[!ht]
	\SetKwData{xtrain}{xtrain} \SetKwData{ytrain}{ytrain} \SetKwData{terminalRF}{terminalRF} \SetKwData{SegMat}{Segmented Matrix} \SetKwData{Temp}{Temporary Matrix} \SetKwData{Repre}{nTreeClus\_Rep} \SetKwData{D}{Dataset}	\KwData{\SegMat of size ($(\mathscr{N}\!\times\!(\mathscr{L}\!-\!n))\times(n+2)$), number of trees ($t$) \textit{with default value of 10}.}
	\KwResult{nTreeClus representation of dataset $D$ (\Repre)}
	\setcounter{AlgoLine}{0}
	initialization;\\
	\xtrain $\leftarrow$ \SegMat[:,1:n]\\
	\ytrain $\leftarrow$ \SegMat[:,n+1] 
	\BlankLine 
	Train RandomForest (X=\xtrain, Y=\ytrain, ntree = $t$)
	
	\terminalRF $\leftarrow$ trace the path of each row in \SegMat to the terminal node for all $t$ trees and store it.
	
	\Repre = An empty DataFrame whose number of columns is equal to \terminalRF's one and whose number of rows is equal to \D's one.
	
	\For{$i \in{\{1,..,\mathscr{N}\}}$}{
		\Repre[i,:]	= $\sum \limits_{j=1}^{(\mathscr{N}\times(\mathscr{L}-n))} \terminalRF[j,:]$ iff \SegMat[j,`y']==i	
	}
	\caption{nTreeClus Representation}
	\label{alg:nTreeClusRepresentation}
\end{algorithm}

To the best of our knowledge, nTreeClus first uses tree-based learning algorithms to encode categorical sequences. Although \citet{1nnBaydogan2016} employed tree-encoder in their Learned Pattern Similarity (LPS) for numeric time series, nTreeClus for the first time generalizes tree-based encoding to sequences. The nature of the categorical sequences is substantially different in that they are not time reliant and are constrained to few levels. Moreover, Unlike the LPS that utilized regression trees, we adopt classification trees. Finally, we introduce a location-based pattern recognizer by adding the position of each substring, depicted in Figure~\ref{fig:seg_seq}.

\subsection{Similarity measure}
The numeric representation generated by nTreeClus is a proper feed to find the distance between each pair of strings. There are numerous ways to find (dis)similarity between these sequences, from which we have implemented Cosine and Manhattan distance to find out which one is the optimal method to be used. We employ hierarchical clustering using Ward's method.

Given two sets of numeric vectors of the same length ($l$), here nTreeClus representations, their Cosine dissimilarity ($\theta$) can be shown as
\begin{equation}
cos(\theta) = 1 - \frac{a.b}{||a||_2 ||b||_2} 
\end{equation}
where $||*||_2$ indicates 2-norm of its argument *, and $a.b$ is the dot product of $a$ and $b$. In positive space, Cosine dissimilarity ranges from 0 to +1, where 0 indicates significant similarity, and +1 declares strong dissimilarity. On the other hand, Manhattan distance places more emphasis on differences between elements of two different vectors. Manhattan distance, or the Minkowski distance with 1-norm, is equal to $||a-b||_1$. The similarity measure selection is relevant to the data types. Therefore, we investigated Manhattan distance, Euclidean distance, and cosine distance in our preliminary experiments. We found that under different sequence characteristics, the cosine distance outperforms other alternatives. This observation is consistent with previous works that emphasized the particular use of the cosine similarity where we have sparse vectors as it saves on computation by considering only non-zero elements~\citep{Li2013, sidorov2014, li2019}.

\subsection{Algorithmic complexity}
The complexity of nTreeClus is comprised of the complexity of two phases, which is $\mathscr{N} \times{(\mathscr{L}-n)}$ for segmentation part and the complexity of Random Forest $O(\sqrt{(\mathscr{L}-n)} \mathscr{N}\log{\mathscr{N}})$; therefore, its complexity depends on the length and the number of sequences. Since tree ensembles can be parallelized, the method has a satisfactory running time. On the other hand, the computational complexity of $k$-mer is $O(\mathscr{N}^2\mathscr{L})$ and quadratically increases if the number of sequences increases~\cite{MSPKmerCounter,edgar2004muscle}. The same case exists for Jaro-Winkler where the time complexity is quadratic $O(\mathscr{L}_p+\mathscr{L}^2)$~\cite{JarowinklerCOMP}. Furthermore, in Levenshtein distance, the time complexity is $O(\mathscr{L}^2)$~\cite{LevenshteinDist}, and similarly, it has a quadratic form. The mixture of HMM has a quadratic relation to the number of hidden states, while both the length and the number of sequences have a direct impact on its complexity.

\section{Experiments and results}
\label{sec:experiments}
To have a fair comparison between nTreeClus and other methods, we define different scenarios imitating the properties of categorical/nominal sequential data with varying levels of intricacy. The algorithm has been applied to the generated datasets with different amounts of $n$ (window size). We conduct a comparison with alternative approaches--e.g., \textit{Similarity-based}, \textit{Feature-based}, and \textit{Model-based Approach}--to make deductions about the competitiveness of nTreeClus. Furthermore, we make the Python implementation of nTreeClus publicly available\footnote{\url{https://github.com/HadiJahanshahi/nTreeClus}}, promoting a full reproducibility of the method for further research. ~\nocite{github}

\subsection{Performance metric}
Clustering is an unsupervised learning approach that determines the innate pattern in a set of unlabeled data while considering the intra-cluster quality and inter-cluster separation~\cite{internalvsexternal}. Two types of cluster validity, namely internal and external criteria~\cite{Clustervaliditymethods}, have been introduced. External validation is the one that is based on pre-specified structure or previous knowledge of data (knowing the ground truth); however, internal validation evaluates the clustering result using features inherited in the dataset when no external criteria are available. In most real cases, using external validation criteria is not feasible; nevertheless, when synthetic data is generated with the known objective functions, associating cluster identifiers with the generated dataset is possible.

In this paper, we use external validation indices whenever we are conversant with the structure of data and adopt internal validation indices whenever we desire to determine the correct number of clusters. 

\subsubsection{Internal cluster validation indices}  

The Calinski-Harabasz Index (CH) is referred to as two top performers for determining the number of clusters~\cite{clusteranalysiseveritt}, representing a global criterion. \citet{CalinskiHarabasz} suggest ``the best sum of squares split'' of the dendrite is where not only is the within-group (cluster) sum of squares (WGSS) minimum, but also the between-group sum of square (BGSS) becomes maximum, corresponding to the maximum value of $CH(k)$ defined as

\begin{equation}
CH(k) = \frac{BGSS}{k-1}\Big/\frac{WGSS}{n-k}
\end{equation}
where $n$ is the number of clustered samples, and $k$ is the number of clusters. On the other hand, the average silhouette width (ASW)~\cite{SilhouettesROUSSEEUW}, given a partition $\xi_{S}$, is defined as
 
\begin{equation}
ASW = \frac{1}{N}\sum\limits_{i=1}^{N}(s(i))
\end{equation}
where $s(i)$, silhouette width, is obtained by 

\begin{equation}
s(i) = \frac{b(i)-a(i)}{max\{a(i),b(i)\}}
\end{equation}
with $a(i)$ being the average dissimilarity of the object $i$ to all other objects of $A$, the cluster to which it has been assigned, and if $d(i,C)$ is defined as the average dissimilarity of the object $i$ to all objects of $C$, any cluster that is different from $A$, then $b(i) = \underset{C\neq A} {minimum} \ d(i,C)$.

$s(i)$ takes values in $[-1,+1]$, where +1 indicates the instance is ``well-classified'', -1 means it is ``misclassified'', and zero denotes it is unclear whether it should be assigned to its cluster or the neighboring cluster.

As the last internal validity index, the Dunn Index~\cite{Dunindex2} measures the ratio between the smallest intercluster and the greatest intra-cluster distance. Let $\delta(C_q,C_r)$ be the intercluster distance metric between clusters $C_q$ and $C_r$, then for the number of clusters $k$, the Dunn Index is
\begin{equation}
DI_k = \underset{1\leqslant{q}\leqslant{k}} {min} \ \bigg\{ \underset{r\neq q}{\underset{1\leqslant{r}\leqslant{k}} {min}} \delta(C_q,C_r) \Bigg/  \underset{1\leqslant{p}\leqslant{k}} {max} \  \\  \Delta(C_p)\\  \bigg\}
\end{equation}
where $\Delta(C_p)$ is the diameter of cluster $C_p$ or the maximum intra-cluster distance. The larger the value of $DI_k$, the more exact the clustering solution.  

\subsubsection{External cluster validation indices}  
When a priori information of the dataset is available, external cluster validation indices can be used. A recent study~\cite{SowmiyaValarmathi}, considering most of the papers in the field of categorical time series, suggests that Clustering Accuracy/Purity, Rand Index (RI)/Adjusted Rand Index (ARI), and F-measure are the most common External Validation measures. We incorporate those methods in the simulation part wherever the ground truth is known. 

Given a set of $n$ elements $S=\{X_1,...,X_n\}$, and their clusters, RI is the proportion of pairs of the two clusters that either belongs to the same cluster or different clusters. RI~
ranges from 0 (with no similarity) to 1 (the same). The Adjusted Rand Index (ARI)~\cite{AdjustedRandIndex} is an index-corrected-for-chance version of RI bounded above by 1--i.e., perfect agreement-- and takes on the value 0 when partitions are selected randomly. Cluster Purity, as one of the frequently used external measures~\cite{DataClustering}, determines to which extent clusters contain a single class. Given $P_j = \frac{1}{n_j}Max_i\Big(n_j^i\Big)$ as the purity of cluster $j$ with class label $i$, the overall Purity is defined as

\begin{equation}
Purity = \sum_{j=1}^k \frac{n_j}{n}Pj,
\end{equation} 
which is the weighted sum of the individual cluster's purity. Here, $k$ is the number of clusters, $n_j$ is the size of cluster $j$, and $n$ is the total number of objects. High values of Purity are desirable. Finally, the F-measure, the harmonic mean of the precision and recall, can be used as an external cluster criterion that takes the values within $[0,1]$, where a higher value indicates a better clustering performance. F-measure is a useful method to evaluate clustering structure~\cite{Handbook}. 
 
\subsubsection{Comparing with Nearest Neighbor classifier}  
The K-NN classifier presumes that the classification of instances can be done by associating the unknown object to the known using the given dis(similarity) function. To measure the efficacy of methods using ``leave-one-out'', we apply one-nearest-neighbor wherever the ground truth of a data set is available~\cite{1nnBaydogan2016}.

\subsection{Patterns to be identified}
\label{sub:patterns}
\begin{figure}[!ht]
	\centering	\includegraphics[scale=0.3]{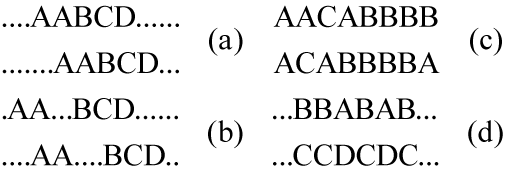}
	\caption{(a): repeated pattern in different positions; (b): repeated gapped pattern; (c): shifted pattern in a closed loop; (d): same pattern with different alphabets.}
	\label{fig:challenges}
\end{figure}

Figure~\ref{fig:challenges} illustrates some of the patterns that may exist in a sequence. Figure~\ref{fig:challenges}(a) depicts a shifted pattern. Figure~\ref{fig:challenges}(b) is a modification of the previous type called gapped-pattern. We show a closed lopped version of the first type in Figure~\ref{fig:challenges}(c). Finally, Figure~\ref{fig:challenges}(d) represents a rare case of sequences where the same structure and order with different alphabets exist. Due to the scarcity of the last case, we only examine the first three patterns and compare all methods in the following experiments. 

\subsection{Sensitivity analysis of the parameter \textit{n}} \label{sec:sensitivity}
The parameter $n$ in nTreeClus is similar to the parameter $k$ in $k$-mers. Using a large $n$ has the advantage of covering a general relation among the elements; by contrast, the small $n$ is superior wherever the pattern is short, and we have noise in data. If the size of the pattern is known and equal to $\tau$, the window size $(n)$ should be set as $\tau$ to capture whole substrings with an equal length; nonetheless, in most actual scenarios, the size of a pattern is unknown. To analyze the sensitivity of the model to the parameter $n$, we compare nTreeClus and $k$-mers ($n$-grams). 

We conduct a simulation including $\mathscr{B}= 40$ batches of sequences. Each batch is formed from $\mathscr{N}= 180$ sequences with different characteristics. They may take $a=\{7, 20\}$ different categorical values in $\mathscr{A}=\{A, B, C, ..., S, T\}$ with the length of $\mathscr{L} = 40$. Eventually, a fixed random pattern with the length of $\mathscr{N}_P= \{8, 15\}$, chosen from the same set $\mathscr{A}$, has been inserted in a random position of half of the sequences. To eliminate any biased randomness, we iterate the simulation ten times. 

\begin{figure}[!ht]
	\centering
	\includegraphics[width=\linewidth]{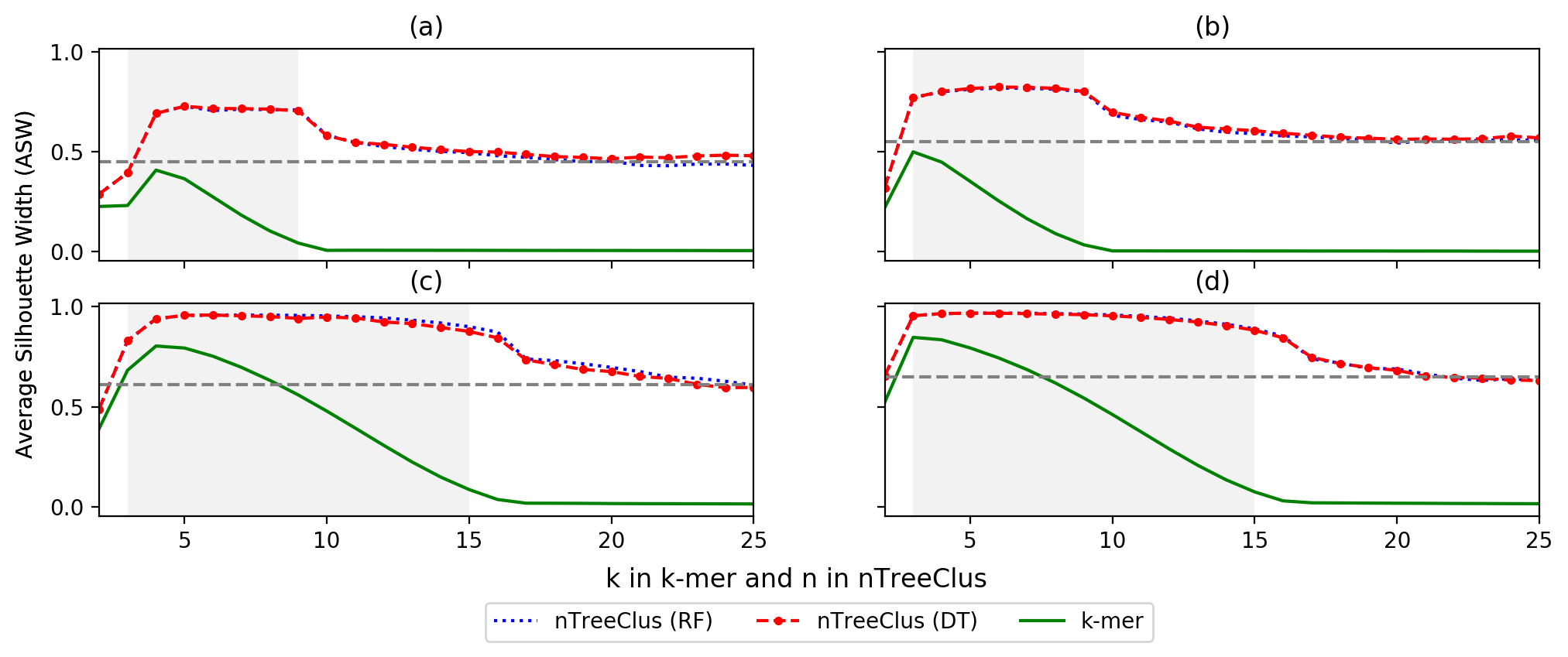}
	\vspace{-1cm}
	\caption{Comparing the robustness of $k$-mer and nTreeClus to their parameters, $k$ and $n$ respectively. We change window size ($k$ or $n$) from 1 to 25 during 10 iterations given in (a): $a = 7$ \& $\mathscr{N}_P= 8$; in (b): $a = 20$ \& $\mathscr{N}_P= 8$; in (c): $a = 7$ \& $\mathscr{N}_P= 15$; and in (d): $a = 20$ \& $\mathscr{N}_P= 15$.}
	\label{fig:windowsize}	
\end{figure}

Figure~\ref{fig:windowsize} presents the result of the experiment on four different datasets with the above properties. In Figures~\ref{fig:windowsize}(a) and \ref{fig:windowsize}(b), the length of the pattern ($\mathscr{L}_P$) is 7 and only the number of alphabets selected from set $\mathscr{A}$ is different. As the number of alphabets increases (going from Figure~\ref{fig:windowsize} (a) to Figure~\ref{fig:windowsize} (b)), so does the Average Silhouette Width (ASW) in that the noise in the sequences increases and recognizing the pattern turns less challenging. The general form of both graphs is almost identical. As long as the $n$ in nTreeClus is less than or equal to the length of pattern (the gray area), the ASW for nTreeClus is stable, and afterward, it converges to a less but acceptable ASW. $k$-mers, by contrast, experiences no stable region, no convergence, and even no specific correlation with the pattern length. In figures~\ref{fig:windowsize}(c) and \ref{fig:windowsize}(d), the length of the pattern ($\mathscr{L}_P$) is larger, and, therefore, it is easier to be recognized. Again, nTreeClus possesses the stable gray region of accuracy (for $3 \leqslant n \leqslant \mathscr{N}_P$), and then asymptotically approaches a reasonable value for ASW. Conversely, $k$-mer lacks the safe region and does not show an acceptable performance for the greater values of $k$.

Figure~\ref{fig:windowsize} also illustrates that for the randomly simulated datasets, nTreeClus consistently outperforms its rival $k$-mers. In this simulation, nTreeClus is robust to classification method selection since both RF and DT lead to almost identical consequences.

The simulation puts forward that if the ground truth is available, the best $n$ is the value close to $\mathscr{L}_P$. For the cases where we have no knowledge about data, the second simulation has been carried out and a default value for parameter $n$ is examined. Three different values for $n$ are inspected, namely square root, one-half, and one-tenth of the length of sequences, $\mathscr{L}$. It showed $n=\sqrt{\mathscr{L}}$ is the optimal $n$ for nTreeClus (see~\ref{sec:best_n}).

\subsection{Simulations}
As described in Section~\ref{sub:patterns}, in this part, we mimic different scenarios that may occur as a pattern in a sequence. The simulated datasets have either typical shifted patterns or more complicated shifted-gapped patterns. Their length and number of alphabets differ case by case.

\subsubsection{Simulation 1 - pattern recognition}
The first simulation is comprised of $\mathscr{B}$=1,080 batches of sequences, each of which is formed from $\mathscr{N}$= \{40, 120, 200\} sequences with different characteristics. They may take $a=\{4, 6, 10, 20\}$ different categorical values in $\mathscr{A}$=\{A, B, C, ..., S, T\}. The length of the sequences ($\mathscr{L}$) is taken from the set of $\{20, 40, 90\}$ while it is fixed for each batch. Eventually, a random pattern with the length of $\mathscr{L}_P$= \{6, 10, 15\} has been inserted in half of the sequences of each batch, thus generating two clusters ($C = 2$). To eliminate any biased randomness, we repeat the simulation ten times. The ultimate output of these replications consists of $\mathscr{B}_S$=129,600 sequences.  

Table~\ref{table:simu1} demonstrates the results of the six methods for the first scenario. It indicates nTreeClus, whether it uses Decision Tree or Random Forest, leads to substantially higher indices than those of the other algorithms. The closest approach to nTreeClus is $k$-mers for $k=\sqrt{\mathscr{L}_i}$ whose Average Silhouette Width (Internal Validation) is half of that of nTreeClus. Nonetheless, we expected low internal validation for all methods because many random noises surround a small pattern in each sequence. Another somewhat surprising result is the robustness of nTreeClus to the method selection, whether it is DT or RF. In MHMM, ranked third based on External indices, the number of latent states ($m$) is chosen based on Bayesian Information Criterion (BIC), $BIC = -2l(\theta) + p \log{m}$ for $m$ in $[1,5]$. We choose the same setting for the following simulations as well. Assuming no foreknowledge is available, the prior probability is selected randomly; hence, its accuracy changes constantly.  

\begin{table*}[!ht]
	\centering
	\caption{Clustering goodness in Simulation 1; here, $k$ in $k$-mer and $n$ in nTreeClus is equal to the square root of the length of sequences.}
	\label{table:simu1}
	\resizebox{\textwidth}{!}{
	\begin{tabular}{lrrrrcr}
		\toprule
		\multirow{2}{*}{Methods} &  \multicolumn{4}{c}{External Validation} & Internal Validation & \multirow{2}{*}{1NN} \\
  & Purity   & RI    & ARI   & F-meas & ASW & \\
		\hline
		nTreeClus (DT)  &  0.988 & 0.980 & 0.959  & 0.987 & \textbf{0.655} & 0.986 \\
		nTreeClus (RF)  &  \textbf{0.992} & \textbf{0.984} & \textbf{0.969} & \textbf{0.990} & 0.651 & \textbf{0.989} \\
		nTreeClus (DT)*  &  0.973 & 0.949 & 0.898  & 0.962 & 0.529 & 0.978 \\
		nTreeClus (RF)*  &  0.984 & 0.970 & 0.940  & 0.979 & 0.546 & 0.986 \\
		Levenshtein  &  0.892 & 0.830 & 0.660  & 0.868 & 0.343 & 0.932 \\
		Jaro-Winkler  &  0.654 & 0.497 & -0.006 & 0.508 & 0.007 & 0.234 \\
		$k$-mers &{} &{} &{} &{} &{} &{}\\
		\qquad $k=1$  &  0.793 & 0.688 & 0.377 & 0.762 & 0.385 & 0.740 \\
		\qquad $k=2$  &  0.910 & 0.861 & 0.721 & 0.901 & 0.420 & 0.894 \\
		\qquad $k=3$  &  0.967 & 0.945 & 0.891 & 0.963 & 0.442 & 0.958 \\
		\qquad $k=\sqrt{\mathscr{L}_i}$  &  0.971 & 0.936 & 0.873 & 0.939 & 0.346 & 0.937 \\
		MHMM  & 0.960 & 0.849 & 0.700 & 0.816 & - & - \\
		\bottomrule
		\multicolumn{7}{l}{\footnotesize \quad *: The position of substrings is used in matrix segmentation.} 
	\end{tabular}
}
\end{table*}

\subsubsection{Simulation 2 - gapped pattern recognition}
Unlike the rigid and simple nature of the first simulation, the second one consists of sequence patterns with gaps where the relative distance between patterns is dynamic. Here, patterns $x$ and $y$ with the length of $\mathscr{N}_{(P_x,P_y)}$= \{(2,3), (4,6), (6,9)\}, respectively, have been inserted in each half of the data while the gap in between changes randomly. For $x$=\{A,C\} and $y$=\{A,B,E\}, it is expected to have them in sequences with different gap in between; for instance, ..\textbf{AC}AABDBA\textbf{ABE}.., ..\textbf{ACABE}.., or ..\textbf{AC}BE\textbf{ABE}.. . The other hyperparameters, including $\mathscr{N}$, $a$, $\mathscr{A}$, $\mathscr{L}$, and $\mathscr{B}_S$, are the same as the first simulation. The second simulation with a dynamic gap and shorter lengths of patterns aims to identify intricate patterns in a given categorical sequence.

Table~\ref{table:simu2} indicates that the general performance of the six methods has reduced as the structure of the data becomes more complicated. Once again, it shows that nTreeClus results in considerably higher indices than those of other algorithms. Here, 3-mer and MHMM are the closest opponents of the current method. On account of the greater complexity of this new dataset, the ASW index for all the methods decreases, whereas the external indices still produce a satisfactory result. Again the worst performance corresponds to the Jaro-Winkler distance in that using edit distance is more rational whenever the length of the sequence is short; namely words and names~\cite{christen2006comparison}. 

\begin{table*}[!ht]
	\centering
	\caption{Clustering goodness measures in Simulation 2; $k$ in $k$-mer and $n$ in nTreeClus is equal to the square root of the length of sequences.}
	\label{table:simu2}
	\resizebox{\textwidth}{!}{
	\begin{tabular}{lrrrrcr}
		\hline
		\multirow{2}{*}{Methods} &  \multicolumn{4}{c}{External Validation} & Internal Validation & \multirow{2}{*}{1NN} \\
		& Purity   & RI    & ARI   & F-meas & ASW & \\
		\hline
		nTreeClus (DT)  &  0.939 & 0.885 & 0.771 & 0.904 & \textbf{0.473} & 0.909 \\
		nTreeClus (RF)  &  \textbf{0.943} & \textbf{0.899} & \textbf{0.798} & \textbf{0.916} & 0.466 & \textbf{0.920} \\
		nTreeClus (DT)*  &  0.923 & 0.853 & 0.707  & 0.879 & 0.387 & 0.903 \\
		nTreeClus (RF)*  &  0.935 & 0.881 & 0.763  & 0.902 & 0.378 & 0.919 \\
		Levenshtein  & 0.869 & 0.796 & 0.592 & 0.839 & 0.319 & 0.868 \\
		Jaro-Winkler  &  0.655 & 0.496 & -0.006 & 0.507 & 0.007 & 0.241 \\
		$k$-mers &{} &{} &{} &{} &{} &{}\\
		\qquad $k=1$  &  0.788 & 0.677 & 0.355 & 0.748 & 0.374 & 0.728 \\
		\qquad $k=2$   &  0.888 & 0.825 & 0.650 & 0.870 & 0.368 & 0.864 \\
		\qquad $k=3$   &  0.927 & 0.886 & 0.773 & \textbf{0.917} & 0.335 & 0.907 \\ 
		\qquad $k=\sqrt{\mathscr{L}_i}$   &  0.902 & 0.774 & 0.549 & 0.777 & 0.195  & 0.808 \\
		MHMM  &  \textbf{0.943} & 0.783 & 0.571 & 0.732 & - & - \\
		\bottomrule
		\multicolumn{7}{l}{\footnotesize \quad *: The position of substrings is used in matrix segmentation.}
		\end{tabular}
}
\end{table*}

\subsubsection{Simulation 3 - analyzing sensitivity to pattern location}
The third simulation is comprised of $\mathscr{B}$=360 batches of sequences, each formed from $\mathscr{N}$= \{40, 120, 200\} sequences with different characteristics. They may take $a=\{5,7,11,16\}$ different categorical values in $\mathscr{A}$=\{A, B, C, ..., O, P\}. Here, the length of the sequences ($\mathscr{L}$) oscillates between 80 to 120 and is not fixed even for each batch. Eventually, a random pattern with the length of $\mathscr{L}_P$= \{6,10,20\} chosen from the set $\mathscr{A}$ has been inserted in each half of the sequences of each batch, thus generating two clusters ($C = 2$). Here, we have inserted the same pattern in each batch while the position of the pattern in half the sequences is identical. For example, we added a pattern in the fifth position of half of the data and the same pattern in the twelfth position of the other half. In other words, we defined two clusters with the same pattern but in different locations. 

\begin{table*}[!ht]
	\centering
	\caption{Clustering goodness measures in Simulation 3; here, $n=10$ in nTreeClus is equal to the square root of the average length of sequences.}
	\label{table:simu4}
	\resizebox{\textwidth}{!}{
	\begin{tabular}{lrrrrcr}
		\toprule
		\multirow{2}{*}{Methods} &  \multicolumn{4}{c}{External Validation} & Internal Validation & \multirow{2}{*}{1NN} \\
		& Purity   & RI    & ARI   & F-meas & ASW & \\
		\hline
		nTreeClus (DT)  & 0.740 & 0.526 & 0.054 & 0.538 & 0.024 & 0.523  \\
		nTreeClus (RF)  & 0.779 & 0.525 & 0.053 & 0.522 & 0.014 & 0.534 \\
		nTreeClus (DT)*  & 0.962 & 0.935 & 0.870 & 0.961 & 0.198 & 0.952 \\
		nTreeClus (RF)*  & \textbf{0.988} & \textbf{0.979} & \textbf{0.957} & \textbf{0.987} & 0.142 & \textbf{0.983} \\
		Levenshtein  & \textbf{0.994} & \textbf{0.989} & \textbf{0.979} & \textbf{0.994} & \textbf{0.441} & \textbf{0.989} \\
		Jaro-Winkler  & 0.658 & 0.495 & -0.008 & 0.501 & 0.006 & 0.350 \\
		$k$-mers &{} &{} &{} &{} &{} &{}\\
		\qquad $k=1$  & 0.652 & 0.503 & 0.006 & 0.535 & 0.184 & 0.609 \\
		\qquad $k=2$  & 0.670 & 0.513 & 0.026 & 0.560 & 0.055 & 0.672 \\
		\qquad $k=3$  & 0.740 & 0.600 & 0.200 & 0.679 & 0.020 & 0.707 \\
		\qquad $k=\sqrt{\bar{\mathscr{L}}}$  & 0.951 & 0.902 & 0.804 & 0.940 & 0.013 & 0.900 \\
		MHMM  & 0.883 & 0.497 & 0.002 & 0.370 & - & - \\
		\bottomrule
		\multicolumn{7}{l}{\footnotesize \quad *: The position of substrings is used in matrix segmentation.}
	\end{tabular}
}	
\end{table*}

Table~\ref{table:simu4} indicates that most of the available methods fail in finding the location of the same pattern in a dataset; nonetheless, Levenshtein (edit distance) and a modification of nTreeClus can capture such behavior in data easily. In the position-sensitive nTreeClus, we incorporate the position of each substring as a new feature; therefore, Decision Tree ensembles treat it as a feature while generating the autoregressive tree. It leads to having a better understanding of the location of a pattern. On the other hand, Levenshtein distance records all the edits needed to transform a substring to another. In this case, the location of the pattern is crucial to have fewer edits --i.e., a fewer number of deletions, insertions, and substitutions is required to turn one string into another. Accordingly, we recommend adding the position factor in nTreeClus wherever the location of the pattern is of importance. 

\subsection{Estimating the number of clusters}
In this experiment, we investigate the ability of each nTreeClus in estimating the number of clusters. Average Silhouette Width, Dunn Index, and Calinski and Harabasz are defined as the metrics examining inter- and intra-clusters sum of squares. The data to measure the efficacy of the method in estimating the number of clusters consists of $\mathscr{B}$=216 batches of sequences where $\mathscr{N}$= \{30,100\}, $a$=\{4, 7, 20\}, $\mathscr{L}$=\{50,100\}, and $\mathscr{L}_P$= \{7, 18\}. This time, the number of clusters ($C$) also changes and takes the values of 3, 6, and 10. The trials have been repeated three times, providing a set of 72 batches with different characteristics. 

All three methods have been utilized to estimate the number of clusters in each batch. Table~\ref{table:EstclusNum} shows the distribution of the estimated number of clusters for different internal cluster validation indices. The surprising result is that ASW, Dunn, and CH show robustness under different scenarios. ASW, with an accuracy of 93.5\%, has the highest performance while CH and Dunn index comes second and third with 77.7\% and 72.2\%, respectively. The last two methods underestimate the number of clusters whenever there is a high level of noise. The experiment suggests that, regardless of the data characteristics, nTreeClus has a high intra-cluster and low inter-cluster similarity. Hence, it can cluster the sequential data with a robust, acceptable internal evaluation score.

\begin{table*}[!ht]
	\centering
	\caption{Distribution of the estimated number of clusters (\^{C}) using internal cluster validation indices for 24 different scenarios, each replicated three times}
	\label{table:EstclusNum}
	\resizebox{0.8\textwidth}{!}{
	\begin{tabular}{l *{12}{c} } 
		\toprule
		\multirow{2}{*}{Index} &  \multicolumn{11}{c}{Number of clusters, \^{C}} & \multirow{2}{*}{median} \\
		& 2 & 3 & 4 & 5 & 6 & 7 & 8 & 9 & 10 & 11-15 & 16-20 &  \\
		\hline
		& \multicolumn{12}{l}{      Sim. 1 ($C = 3$)}\\
		ASW & {} & \textbf{69} & {} & {} & {} & {} & {} & {} & {} & {} & 3 & 3 \\
		Dunn & 10 & \textbf{59} & 3 & {} & {} & {} & {} & {} & {} & {} & {} & 3 \\
		CH & 12 & \textbf{60} & {} & {} & {} & {} & {} & {} & {} & {} & {} & 3 \\
		\hline
		& \multicolumn{12}{l}{      Sim. 2 ($C = 6$)}\\
		ASW & {} & {} & {} & {} & \textbf{66} & {} & {} & {} & {} & 1 & 5 & 6 \\
		Dunn & 17 & {} & {} & {} & \textbf{51} & 4 & {} & {} & {} & {} & {} & 6 \\
		CH & 15 & 1 & {} & {} & \textbf{56} & {} & {} & {} & {} & {} & {} & 6 \\
		\hline
		& \multicolumn{12}{l}{      Sim. 3 ($C = 10$)}\\
		ASW & {} & {} & {} & {} & {} & {} & 1 & 1 & \textbf{67} & 1 & 2 & 10 \\
		Dunn & 23 & {} & {} & {} & {} & {} & {} & {} & \textbf{46} & 3 & {} & 10 \\
		CH & 17 & 1 & {} & 1 & {} & {} & {} & 1 & \textbf{52} & {} & {} & 10 \\
  \bottomrule
	\end{tabular}
}
\end{table*}

\begin{figure*}[!ht]
	\centering
	\includegraphics[width=\linewidth]{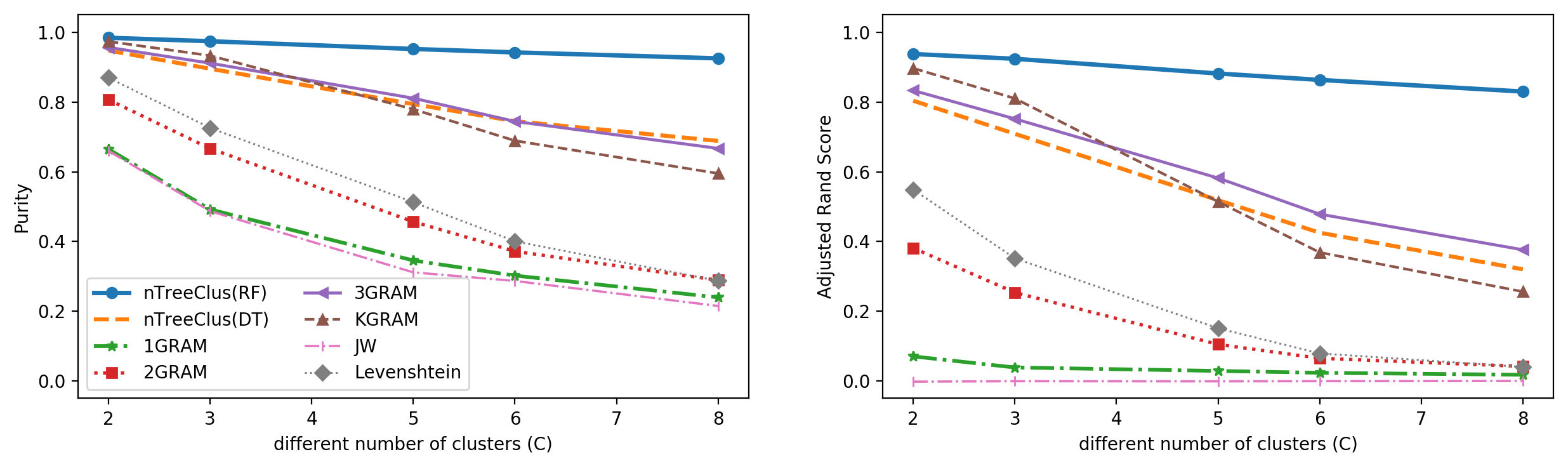}
	\caption{Analyzing robustness/ruggedness of the methods under different amounts of $C$}
	\label{fig:robustness}
\end{figure*}

Figure~\ref{fig:robustness} shows the performance of the methods given a different number of clusters ($C$). The plot indicates that the RF modification of nTreeClus is robust under different values for $C$, whereas the clustering goodness of all other methods sharply decreases as the number of clusters increases. 

\subsection{Real data application}
\textcolor{black}{To validate our method, we apply it to real benchmark datasets of different domains. We employ complete coronavirus genomes and Austrian Wage Mobility data, followed by Nucleotide sequences of transcripts on the reference chromosomes and the protein dataset in~\ref{sec:nucleotide} and~\ref{sec:protein}, respectively.}

\subsubsection{\textcolor{black}{Phylogeny of Coronavirus}} \label{sec:corona}
\textcolor{black}{We utilize a collection of 30 coronavirus genomes ranging in size from 27,000 to 32,000 nucleotides (See \ref{sec:corona_apx} for more details on the dataset). Coronaviruses are enveloped, single-stranded, positive-sense RNA viruses that belong to the Coronaviridae family. They are pleomorphic RNA viruses found in many animals, including bats and humans. This viral family may readily infect new species by crossing the species barrier~\citep{monchatre2017}. Due to the importance of knowing this particular virus after the pandemic, there have been many studies into the classification and evolutionary relationships of these viruses. We leverage our method to examine the entire genome sequences of 30 coronaviruses and four non-coronaviruses as outgroups. According to the host type, the 30 coronavirus sequences are clustered into five categories~\citep{saw2019}. The HCoV-HKU1 virus was thought of as a member of the second group~\citep{woo2005}, whereas \citet{Chenglong2010} later considered it as a separate category, group 5. We follow the same categorization in our work as done in other recent studies~\citep{li2017, monchatre2017}.}

\begin{figure}[!ht]
	\centering
	\includegraphics[width=\linewidth]{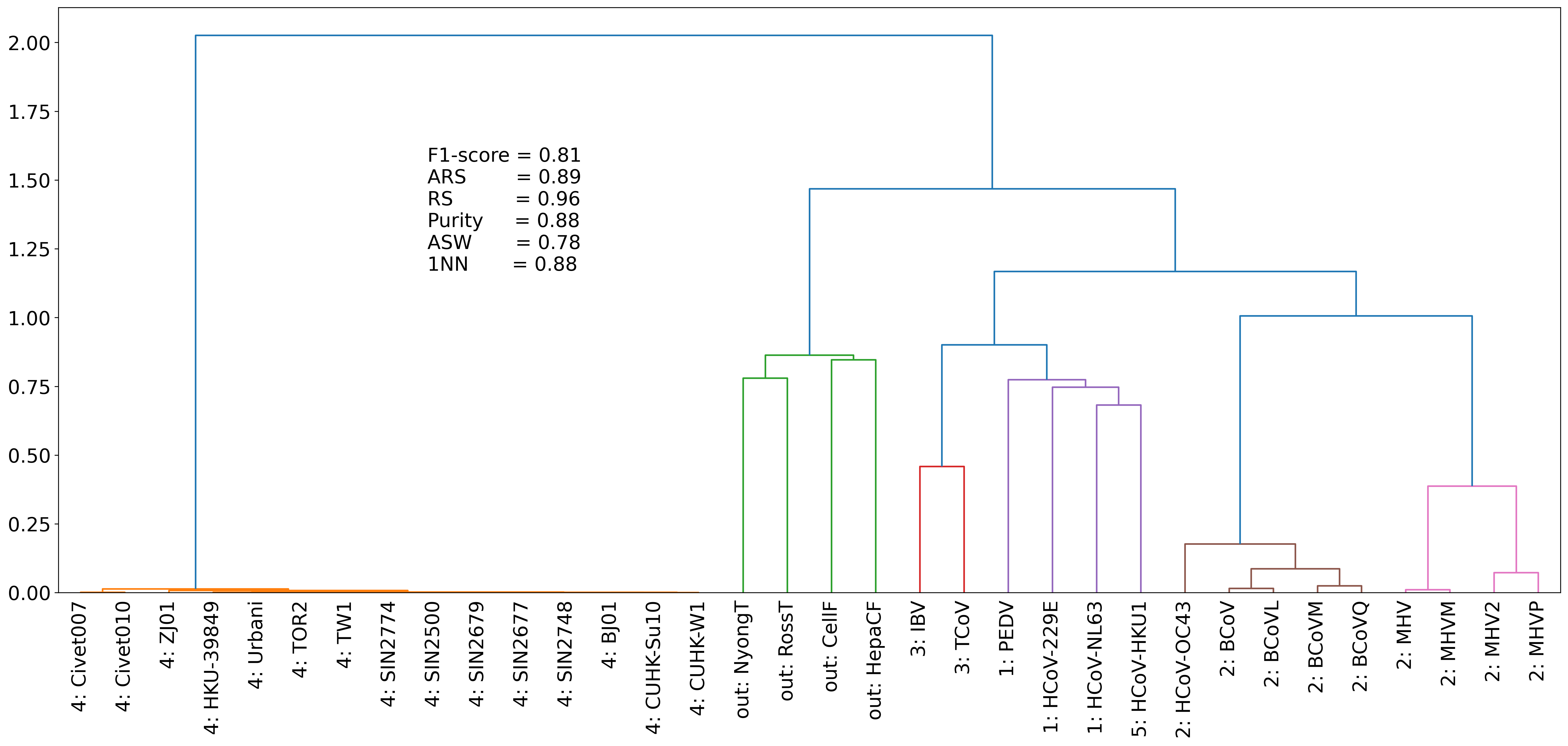}
	\caption{The phylogenetic tree of 30 coronavirus whole genomes based on nTreeClus}
	\label{fig:phylogeny}
\end{figure}

\textcolor{black}{Figure~\ref{fig:phylogeny} shows the phylogenetic tree of coronavirus genomes and outgroup genomes. Accordingly, SARS coronavirus genomes, cluster 4, are tightly clustered and are separated from other clusters. Moreover, the proposed nTreeClus method correctly separates outgroup subsamples from others. It also recommends two subclusters for cluster 2, which is aligned with previous studies~\cite{monchatre2017,li2017,hoang2015}. The only incorrect clustering occurs in the case of HCoV-HKU1, which should be considered as a separate cluster. Our method identifies a high similarity of this genome with those in cluster 1. This finding is different than that of ~\citet{li2017} and \citet{hoang2015} and more similar to the phylogenetic tree of \citet{monchatre2017} (See Figure~\ref{fig:coronavirus_old_paper}). Thus, we recommend further investigation to find the cluster of the controversial HCoV-HKU1 virus. Regarding the performance of our model, we observe a better clustering of the first cluster compared to that of \citet{monchatre2017} reported in \ref{sec:corona_apx}. They showed that their method has advantages over $k$–mer and FFP methods, whereas nTreeClus demonstrates a better performance than theirs in  clustering coronavirus genomes. Hence, we conclude that our model can address real-world problems even with lengthy nucleotides (more than 30,000 in length). }

\subsubsection{Austrian Wage Mobility data}
\label{subsec:AustrianWage}
The Austrian Wage dataset, first used by~\citet{pamminger2010}, reports the wage category in successive years for the young men entering the labor market from 1975 to 1980. It consists of $\mathscr{N}=9402$ of such workers whose gross monthly wages of successive years (ranging from $\mathscr{L} = 2$ to $\mathscr{L} = 32$ years) have been reported. The gross monthly wages are divided into six categories, from 0 to 5, in which zero corresponds to unemployment and categories one to five correspond to the quintiles of the income distribution
. 
 
\begin{figure}[!ht]
	\centering
	\includegraphics[width=0.5\linewidth]{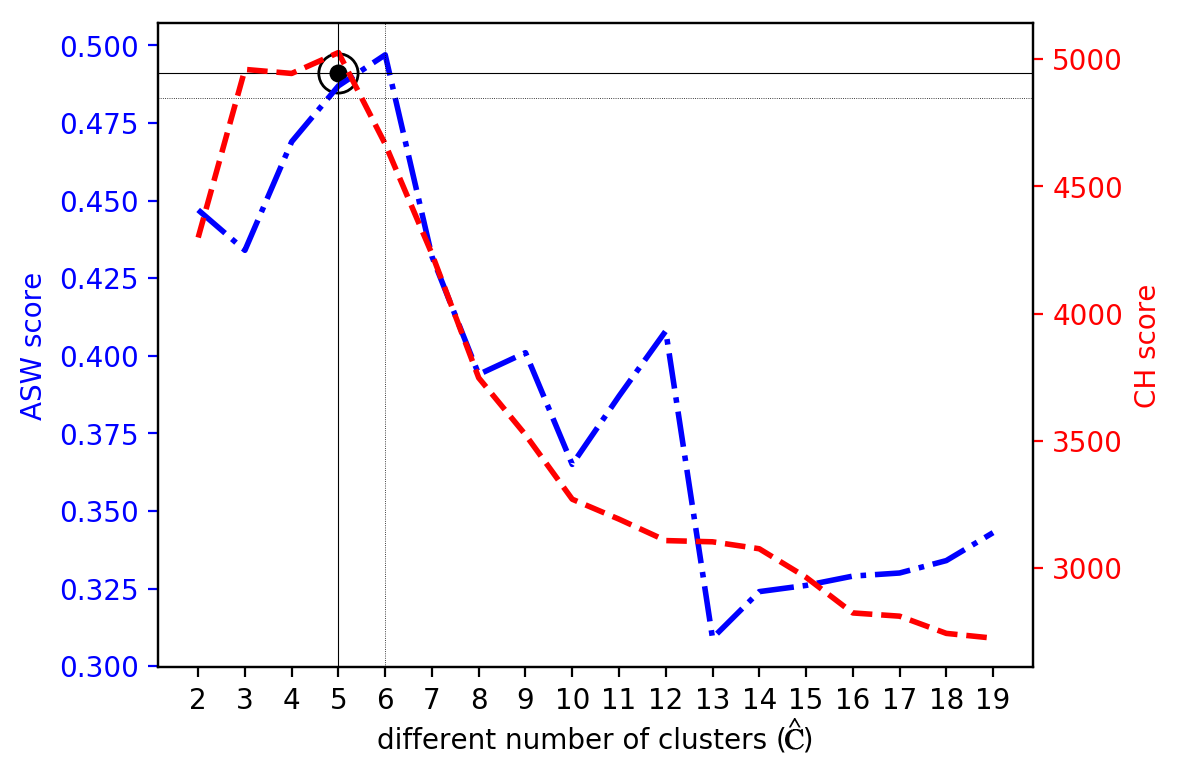}
	\caption{Finding the best number of clusters (\^{C}) using Average Silhouette Width and Calinski-Harabasz Index}
	\label{fig:EmployeeEstNumClus-k2}
\end{figure}

\citet{pamminger2010} use Dirichlet multinomial clustering and Markov Chain clustering, suggesting that the number of clusters is between four and six. We implement nTreeClus on the data, and for the Internal Validations, ASW and CH, it recommends 5 and 6 clusters, respectively. Considering the weighted average score of both validation methods, we determine the number of clusters (\^{C}) as 5 (Figure~\ref{fig:EmployeeEstNumClus-k2}). 

Since the same dataset was examined by~\citet{Garcia-Magarinos2015} with special attention to \^{C} = 4 and \^{C} = 5, and by~\citet{pamminger2010} with special attention to \^{C} = 4, we explore nTreeClus for both scenarios to contrast it with the result of the previous studies. 

Parameter $n$ in nTreeClus is set to 1 since the shortest sequence in the given dataset has a length of 2 years. When $n$ is too small, nTreeClus cannot capture the relation between the categorical time series for the upcoming years. On the other hand, if $n$ increases, short sequences should be inevitably removed from the dataset. To keep the consistency between nTreeClus and the other two methods, we decided not to increase the value of $n$. Despite the compulsion to choose the least possible value for $n$, the output seems easily interpretable. Previously, the dataset has been categorized as \textit{unemployed}, \textit{climbers}, \textit{low-wage}, and \textit{flexible}, while nTreeClus suggests a new behavior in the dataset (shown in the first row of Figure~\ref{fig:4Clusters}), in which flexible (\textit{temporary workers}), climbers to the level 4 of wage (\textit{middle-level employees}), top-level employees (\textit{top managers}), and finally \textit{the low-paid workers} can be seen. Making inferences clearer, the first mixed behavior can be considered as the ordinary workers that have no opportunity to take middle or high-rank positions in the organization, and after a while, they become unemployed. The second group includes the middle-level employees with the chance of attaining a better paying position but not the highest. The third cluster mostly contains top-level employees who move mainly between the fourth and fifth income level. Finally, the last cluster incorporates the ones who constantly work without any improvement in their income, keeping them as low-wage employees during their whole service to the organization. Comparing to Fig. 6 of~\citet{pamminger2010} and Fig. 11 of~\citet{Garcia-Magarinos2015}, nTreeClus includes the \textit{low-wage} and \textit{climbers} while defines new easily interpretable groups of \textit{top managers} and \textit{temporary workers}.

For \^{C} = 5 (Figure~\ref{fig:5Clusters}), suggested by~\citet{Garcia-Magarinos2015} and considered as the optimal number of clusters by nTreeClus, flexible (\textit{temporary workers}) cluster has been split up into two groups. First, the employees who can maintain their position in the organization without quitting the job and even with a limited prospect for promotions, and second, the ones who quit the organization after a short period and remain unemployed. The transition diagram in the second row of Figure~\ref{fig:5Clusters} indicates how each cluster moves among different wage levels. For cluster 1 (the most left Figure), there is a chance of promotion from level 1 to level 3, whereas, for cluster 2, the probability of quitting the job is more than any other state. For cluster 3, going from state three to four, for cluster 4, promoting from state four to five, and for the last cluster, remaining as a low-paid worker (level 1 forever) have the highest likelihood. 

\begin{landscape}
	\begin{figure}
		\begin{center}
			\includegraphics[width=0.85\linewidth]{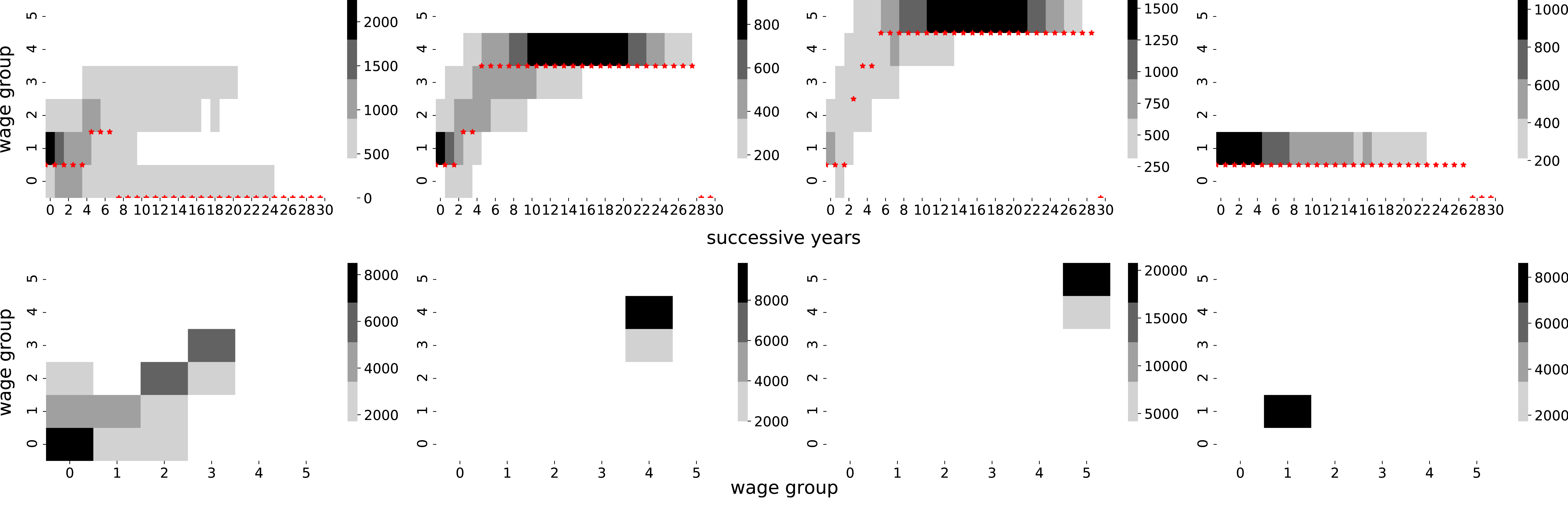}
			\caption{Heat-map of the wage clusters (first row) and the transition between different wage groups (second row) for $\hat{C} = 4$}
		\label{fig:4Clusters}
		\end{center}
	\end{figure}

	\begin{figure}
	\begin{center}
		\includegraphics[width=0.85\linewidth]{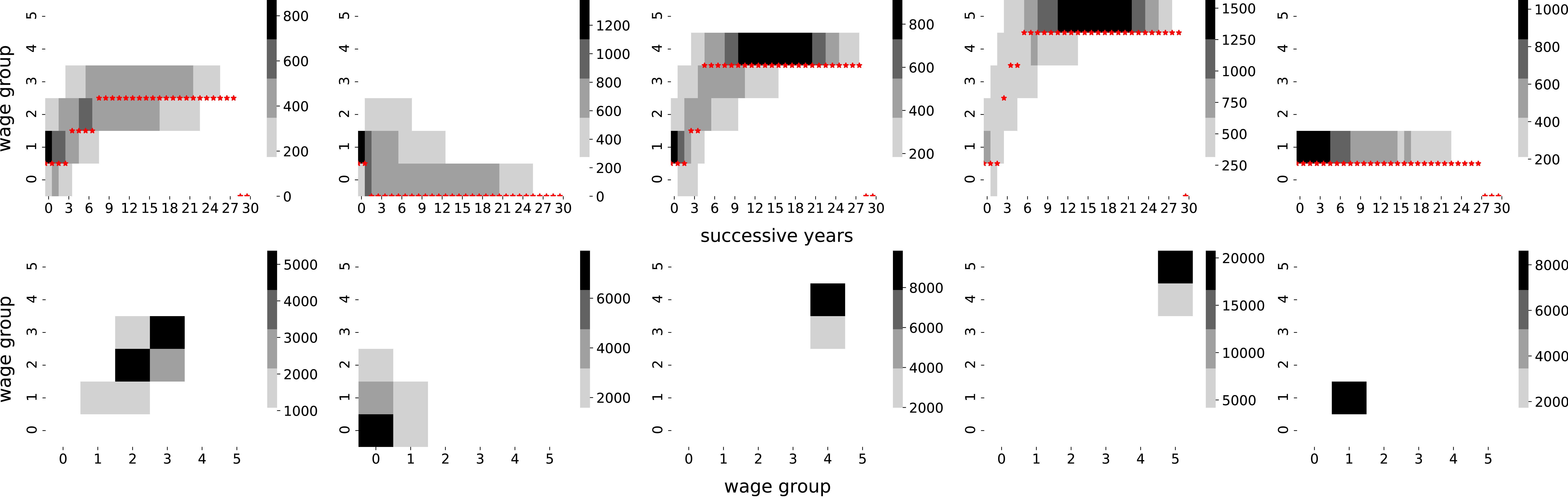}
		\caption{Heat-map of the wage clusters (first row) and the transition between different wage groups (second row) for $\hat{C} = 5$}
		\label{fig:5Clusters}
	\end{center}
	\end{figure}
\end{landscape}

\section{Discussions and future work} \label{sec:discussion}
The results given the diverse datasets we used were promising. nTreeClus shows a high performance given both synthetic and real datasets and demonstrates robustness towards its hyperparameter. Nevertheless, some threats to the validity of our study should be addressed. Wherever a random set is generated, we repeat the experiment ten times to alleviate the likelihood of bias in our report. Furthermore, we consider internal and external performance metrics to ensure that our model captures the pattern in a given sequence. We use cosine similarity in positive space where the expected output is between 0 and 1. Particularly, the cosine similarity is beneficial where we have high-dimensional sparse vectors since only non-zero values should be considered~\citep{li2019}. However, we also used other approaches, e.g., Manhattan distance, to investigate the effect of similarity method selection. We found that in the case of string similarity, given the datasets described in the paper, the cosine similarity outperforms other potential similarity approaches.

A relevant venue for future research would be to explore the method performance for the classification task. Furthermore, nTreeClus should be extended to empirically and theoretically investigate categorical sequences that are continuous series of elements and where the time between two elements matters--e.g., clickstream datasets. The current version of nTreeClus can handle those datasets only through discretizing the time and repeating each element. Also, we plan to extend the study to use path-encoding instead of terminal-node-encoding of the DT algorithm. Indeed, the path that a decision takes to arrive at its terminal node can be traced back and vectorized based on the frequency of visiting each intermediate node. This version increases both the complexity of the calculation and the dimension of the representative vector. However, it captures more information on the autoregressive behavior of sequences.

\section{Concluding remarks} 
\label{sec:conclusion}
In this paper, we argued that sequence mining is a paramount need to be addressed by data scientists and that the state-of-the-art algorithms face major hurdles--e.g., universal applicability, computational complexity, sensitivity to the position of a pattern and the length of sequences, and vulnerability to parameter setting. We further showed empirically that based on high external cluster validation indices and low internal cluster validation indices, those methods overfit on sequences where we are unacquainted with the dataset structure.

To mitigate these potential problems, we introduced a new framework for encoding categorical time series. nTreeClus takes advantage of the current methods, including Tree-based classifiers, $k$-mers, and autoregressive models for categorical sequences.  Although it might seem that nTreeClus relies heavily on existing methods, to the best of our knowledge, this potential of tree-based representation has not been suggested by any researcher. Therefore, the methodology is novel regardless of its fundamental algorithms. nTreeClus proposes a numeric representation for categorical sequences, established upon the autoregressive behavior of the data. In the first phase, we segment sequences into length-smoothed substrings based on the predefined window size. This segmented matrix is a feed for autoregressive Tree ensembles where each rule in a tree describes a pattern in the sequence. In the second phase, nTreeClus encodes sequences by counting the number of observations in each terminal node. It indicates the behavior of sequences based on the correlative structure of DT's rules. Eventually, using similarity methods, we can cluster sequences given the new encoder algorithm. 

Our experimental results show that nTreeClus is robust towards parameter setting and provides computationally efficient and superior outcomes on real and synthetic benchmark datasets with heterogeneous characteristics. We observe low internal validation for all methods where the pattern size was small compared to the sequence size. This fact is inevitable in arbitrary, synthetic pattern generation and can not be considered as the Achilles heel of any method. In the end, we acknowledge the need for applying nTreeClus on sequential datasets of different domains to further investigate its generalizability.

\section*{\textcolor{black}{Acknowledgement}}
\textcolor{black}{The authors would like to thank Mohamed Abuelanin for his help with the Bioinformatics data acquisition and explanation. This research is supported by Air Force Office of Scientific Research Grant [Grant number: FA9550-17-1-0138].}

\bibliographystyle{elsarticle-num-names}
\singlespacing
\bibliography{references}

\appendix
\section{Finding the default value for the parameter $n$}\label{sec:best_n}
\begin{figure}[!ht]
	\centering	\includegraphics[scale=0.55]{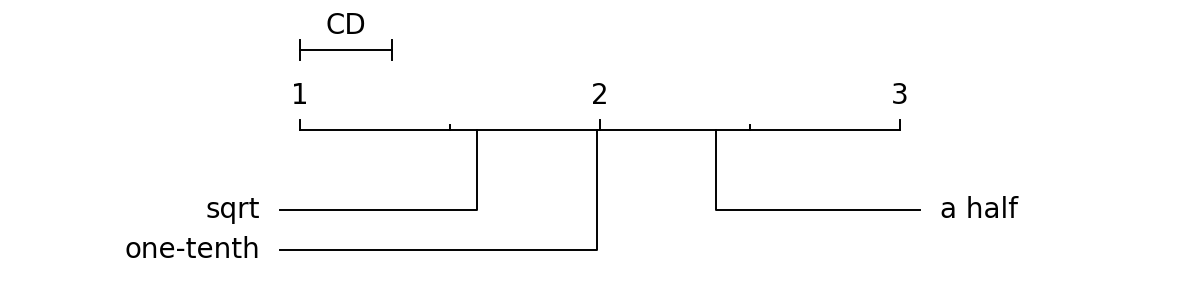}
	\caption{The average rank for three different setting of $n$. The critical difference at significance levels $\alpha = 0.05$ and $= 0.1$ is 0.349 and 0.306, respectively. The performance of $n = \sqrt{\mathscr{L}}$ is statistically better than the others at different $\alpha$ levels.}
	\label{fig:Nemenyi}
\end{figure}

In order to find the optimal value for the hyperparameter of the model, we ran a simulation including 90 batches of sequences with the structure of $\mathscr{N}= \{180, 450, 720\}$, $a=7$ different categorical values in set of $\mathscr{A}=\{A, B, C, ..., G\}$, $\mathscr{N}_P= \{7, 15\}$ , and finally $C = \{2, 5, 8\}$ number of clusters. Unlike the first simulation, the number of clusters varies in order to cover different circumstances that may occur in real cases. After five replications, 40500 sequences with the specified characteristics have been simulated. Figure~\ref{fig:Nemenyi} shows the average ranks for all different settings of $n$. The Friedman test shows a significant difference between the 3 settings at the both 0.05 and 0.1 significance level. Proceeding with the Nemenyi test, Critical Difference (CD) is computed, showing the performance of $n = \sqrt{\mathscr{L}}$ is significantly better than $n = 0.1\mathscr{L}$ and $n = 0.5\mathscr{L}$ at different $\alpha$ levels. Therefore, in the following experiments, $n$ always has been set to the square root of $\mathscr{L}$.

\section{Simulation 4 - length sensitivity analysis}\label{sec:len_sens}

In most real cases, the length of sequences to be compared with each other is varying. In such a case, the method which is robust to the length of sequences can yield the best result. All the properties of the generated sequences, including $\mathscr{B}$, $\mathscr{N}$, $a$, $\mathscr{A}$, $C$, $\mathscr{L}$, and $\mathscr{L}_P$, are the same as simulation 3. To eliminate any biased randomness, we repeat the experiment ten times. Because of ever-changing nature of sequences, parameter $n$ in nTreeClus has been set to $10$ which is equal to the square root of average length $100$ ($n = \sqrt{average(\mathscr{L})} = 10$). 

\begin{table*}[!ht]
	\centering
	\caption{Clustering goodness measures in Simulation 4; here, $n=10$ in nTreeClus is equal to the square root of the average length of sequences.}
	\label{table:simu3}
	\resizebox{\textwidth}{!}{
	\begin{tabular}{lrrrrcr}
	\toprule
		\multirow{2}{*}{Methods} &  \multicolumn{4}{c}{External Validation} & Internal Validation & \multirow{2}{*}{1NN} \\
		& Purity   & RI    & ARI   & F-meas & ASW & \\
		\hline
		nTreeClus (DT)  &  0.971 & 0.951 & 0.901  & 0.969 & \textbf{0.367} & 0.966 \\
		nTreeClus (RF)  &  \textbf{0.983} & \textbf{0.968} & \textbf{0.936}  & \textbf{0.981} & 0.360 & \textbf{0.981} \\
		nTreeClus (DT)*  &  0.964 & 0.940 & 0.880  & 0.962 & 0.334 & 0.959 \\
		nTreeClus (RF)*  &  0.978 & 0.960 & 0.921  & 0.975 & 0.322 & 0.975 \\
		Levenshtein  &  0.815 & 0.748 & 0.496  & 0.787 & 0.146 & 0.898 \\
		Jaro-Winkler  &  0.648 & 0.496 & -0.008 & 0.503 & 0.007 & 0.276 \\
		$k$-mers &{} &{} &{} &{} &{} &{}\\
		\qquad $k=1$  &  0.689 & 0.565 & 0.129  & 0.636 & 0.228 & 0.706 \\
		\qquad $k=2$  &  0.898 & 0.830 & 0.660  & 0.888 & 0.201 & 0.884 \\
		\qquad $k=3$  &  0.969 & 0.946 & 0.892  & 0.967 & 0.228 & 0.961 \\
		\qquad $k=\sqrt{\bar{\mathscr{L}}}$  &  \textbf{0.988} & \textbf{0.977} & \textbf{0.953}  & \textbf{0.988} & 0.087 & \textbf{0.983} \\
		MHMM  &  0.944 & 0.781 & 0.566 & 0.741 & - & - \\
		\bottomrule
        \multicolumn{7}{c}{* the methods in which the position of substrings is used in matrix segmentation.}
	\end{tabular}
}	
\end{table*}

Table~\ref{table:simu3} clearly shows that, based on External Validation indices, both nTreeClus and $k$-mer outperform the others. Notwithstanding the good performance of $k$-mer in External Validation, it is nTreeClus that outperforms all other methods regarding Average Silhouette Width index. Once more, the worst performance corresponds to the Jaro-Winkler and Levenshtein method.

\section{\textcolor{black}{Clustering Nucleotide Sequences}} \label{sec:nucleotide}
\textcolor{black}{We examine the performance of the nTreeClus on Nucleotide sequences extracted from 
GENCODE human annotation version 39\footnote{\href{https://www.gencodegenes.org/human/release_39.html}{gencodegenes.org/human/release\_39.html}}. We download the transcript sequences FASTA file containing ``Nucleotide sequences of all transcripts on the reference chromosomes''. We filter the dataset on the top-10 most frequent gene ids in the dataset, leaving us with 1,728 sequences whose lengths vary from 144 to 24,124. The dataset with its 10 clusters is available on our GitHub page\footnote{\href{https://github.com/HadiJahanshahi/nTreeClus}{github.com/HadiJahanshahi/nTreeClus}}.
}

\textcolor{black}{Table~\ref{table:nucleotide} shows the performance of different versions of nTreeClus and the baseline methods. Our proposed algorithm is capable of handling the clustering task of sequences of different lengths in a short time. We also observe a significantly better performance of nTreeClus in terms of external metrics compared to all other methods. Although $k$-mer shows a slight improvement according to internal indices, it may originate from the false clustering of some instances based on external metrics. Overall, we find decent clustering goodness of the nTreeClus in this real application.}

\begin{table}[!ht]
	\centering
	\caption{\textcolor{black}{Clustering goodness measures in Nucleotide Sequence dataset; here, $n=42$ in nTreeClus is equal to the square root of the average length of sequences.}}
	\label{table:nucleotide}
	\resizebox{\textwidth}{!}{
    \begin{tabular}{lrrrrcr}
    \toprule
        \multirow{2}{*}{Methods} &  \multicolumn{4}{c}{External Validation} & Internal Validation & \multirow{2}{*}{1NN} \\
		\cmidrule(lr){2-5}
		& Purity   & RI    & ARI   & F-meas & ASW & \\
		\hline
        nTreeClus (DT)  & 0.852 & 0.936 & 0.690 & 0.859 & 0.491 & 0.999 \\
        nTreeClus (RF)  & 0.855 & 0.938 & 0.696 & 0.860 & 0.498 & 0.999 \\
        nTreeClus (DT)* & \textbf{0.867} & \textbf{0.945} & \textbf{0.724} & \textbf{0.872} & 0.491 & 0.999 \\
        nTreeClus (RF)* & 0.865 & 0.943 & 0.718 & 0.870 & 0.491 & 0.999 \\
        Levenshtein & 0.620 & 0.653 & 0.084 & 0.188 & 0.411 & 0.972 \\
        Jaro-Winkler & 0.202 & 0.802 & 0.000 & 0.093 & -0.014 & 0.064 \\
        $k$-mers \\
        \qquad $k=1$ & 0.498 & 0.811 & 0.165 & 0.306 & 0.264 & 0.521 \\
        \qquad $k=2$ & 0.554 & 0.810 & 0.202 & 0.366 & 0.199 & 0.894 \\
        \qquad $k=2$ & 0.663 & 0.878 & 0.399 & 0.557 & 0.128 & 0.964 \\
        \qquad $k=\sqrt{\bar{\mathscr{L}}}=n$ & 0.848 & 0.914 & 0.616 & 0.767 & \textbf{0.566} & \textbf{1.000} \\
        \bottomrule
        \multicolumn{7}{c}{* the methods in which the position of substrings is used in matrix segmentation.}
    \end{tabular}
}
\end{table}

\section{Clustering Protein Dataset} \label{sec:protein}
As the fourth real dataset, a protein dataset has been taken from~\cite{paynabar2016sequence}, containing $\mathscr{N} = 2112$ sequences with the length $\mathscr{L}$ between 80 and 127. The protein dataset has $a = 20$ alphabets (amino acids). Each sequence has one of two tasks, viz. ``Might get involved in a Signal Recognition Particle (SRP) pathway'' or ``Bind to DNA and changes its conformation''. Sequences are labeled based on their functions and have a reasonably balanced distribution (979 of the first task and 1133 of the second one).

\begin{table*} [!ht]
	\centering
	\caption{Clustering goodness measures in Protein Dataset; here, $k$ in $k$-mer and $n$ in nTreeClus is equal to the square root of the average length of sequences ($k=n=11$).}
	\label{table:protein}
	\resizebox{0.9\textwidth}{!}{
	\begin{tabular}{l *{5}{c} }
		\hline
		\multirow{2}{*}{Methods} &  \multicolumn{4}{c}{External Validation} & Internal Validation \\
		& Purity   & RI    & ARI   & F-meas & ASW \\
		\hline
		nTreeClus (DT)  &  1.000 & 1.000 & 1.000 & 1.000 & 0.815 \\
		nTreeClus (RF)  &  1.000 & 1.000 & 1.000 & 1.000 & 0.832 \\
		Levenshtein  & 1.000 & 1.000 & 1.000 & 1.000 & 0.950 \\
		Jaro-Winkler  &  0.750 & 0.500 & -0.001 & 0.480 & -0.003 \\
		$k$-mers & 0.949 & 0.903 & 0.805 & 0.948 & 0.444 \\
		MHMM  & 1.000 & 1.000 & 1.000 & 1.000 & - \\
		\bottomrule
	\end{tabular}
}	
\end{table*}

The result shown in Table~\ref{table:protein} demonstrates that nTreeClus has a pure result based on External Validation criteria, while for Internal Validation, it is Levenshtein that has the highest Average Silhouette Width. In previous datasets, Levenshtein obtained a low score indicating that although there is a probability to get a proper result through the method, this result is highly correlated with the type of the pattern in the sequence. Again, nTreeClus is among the best methods based on the quality of clustering (Internal or External Validation indices).

\section{Sensitivity analysis} \label{sec:sensitivity2}
In Section \ref{sec:sensitivity}, we explored the sensitivity of nTreeClus to its parameter $n$. However, we aim to further discuss its sensitivity to dataset characteristics, i.e., the number of instances and the length of sequences. The proposed algorithm is expected to be applicable when we have few or many instances. It also requires to have a decent performance given sequences of different lengths. We conduct two experiments to investigate the sensitivity of the model to data characteristics.

\subsection{Sensitivity to the number of instances}
\begin{figure}[!ht]
    \centering
    \includegraphics[width=0.5\linewidth]{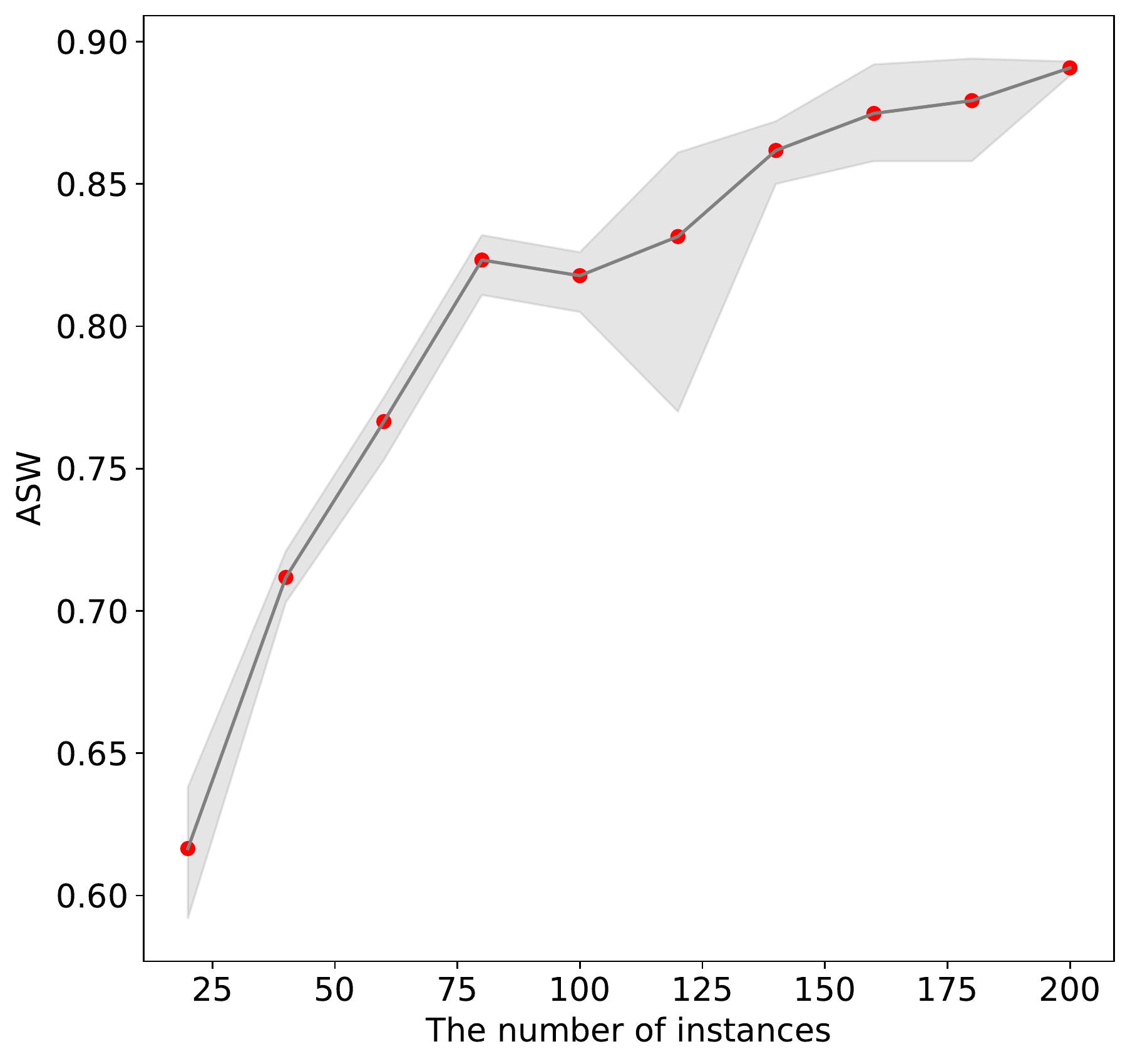}
    \caption{The sensitivity of nTeeClus to the number of instances}
    \label{fig:sensitivity_inst}
\end{figure}

In this experiment, we investigate the sensitivity of the model to the number of instances. Accordingly, we keep different data features constant while changing the number of instances. The dataset characteristics are summarized as  $\mathscr{N}$= \{20, 40, \dots , 200\}, $a$=10, $\mathscr{L}$=40, $\mathscr{L}_P$= 10, and $C = 2$. The trials have been repeated four times, providing a set of 40 batches with different characteristics. Figure~\ref{fig:sensitivity_inst} shows how the average silhouette width changes as the number of instances increases from 20 to 200. It illustrates that as the number of instances increases, the vector representation improves in terms of ASW values. Nevertheless, the model performance given external indices is always 1 regardless of the changes in the number of instances. Having more instances results in clusters with dense intra-cluster distances and farther inter-cluster distances.

\subsection{Sensitivity to sequence lengths}
\begin{figure}[!ht]
    \centering
    \includegraphics[width=0.5\linewidth]{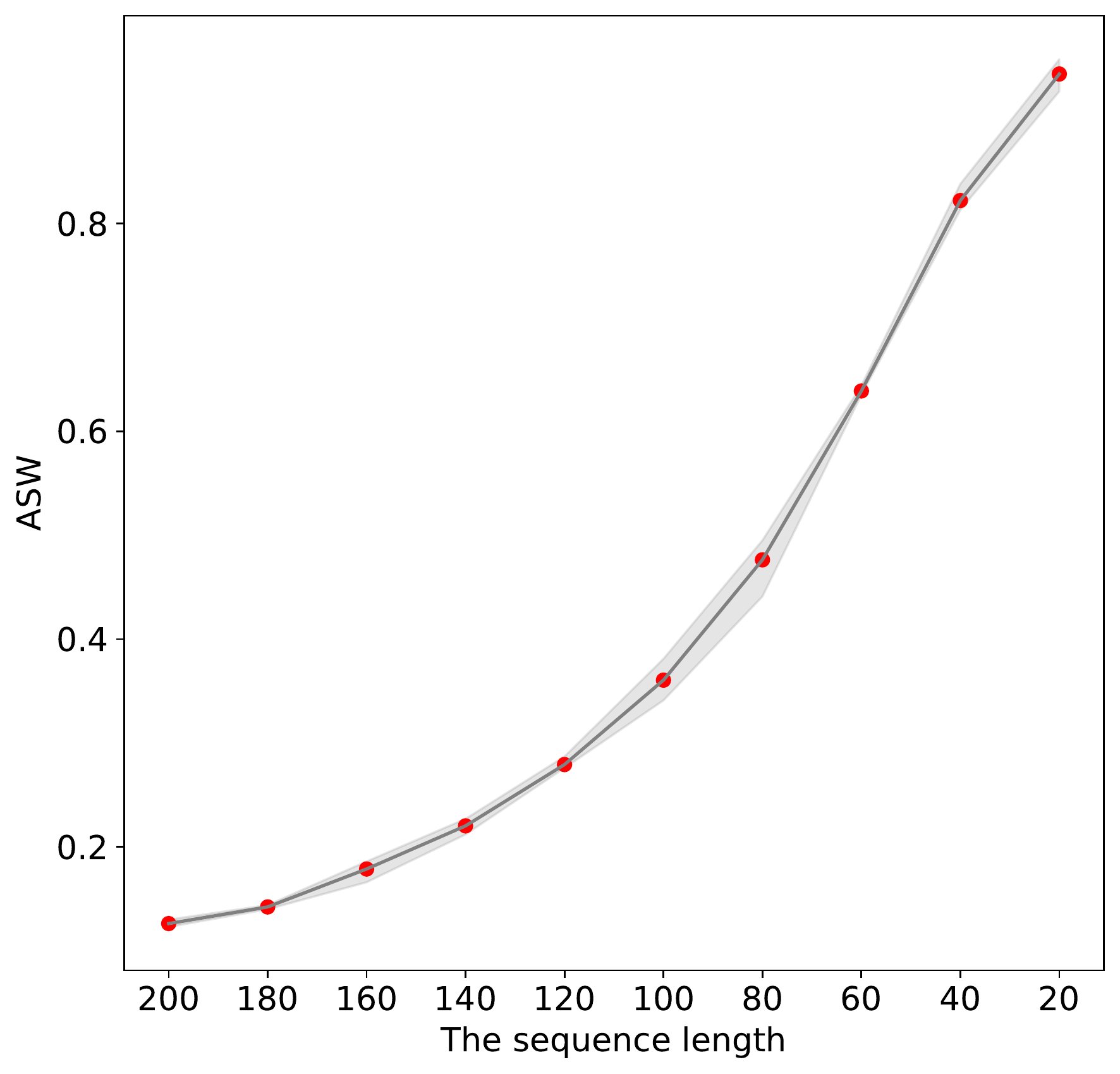}
    \caption{The sensitivity of nTeeClus to the sequence length (The x-axis is in reverse order.)}
    \label{fig:sensitivity_seq_len}
\end{figure}

In the second experiment, we analyse the model sensitivity to the change in sequence lengths while keeping the pattern length constant. We also keep other data features intact. The datasets are featured by  $\mathscr{N}$= 100, $a$=10, $\mathscr{L}$=\{20, 40, \dots , 200\}, $\mathscr{L}_P$= 10, and $C = 2$. The trials have been repeated four times, generating 40 batches with different features. Accordingly, the pattern to sequence length ratio ($\nicefrac{\mathscr{L}_P}{\mathscr{L}}$) changes from 0.5 to 0.05. This fact explores the stability of the model towards short or long sequential patterns. Figure~\ref{fig:sensitivity_seq_len} demonstrates the average silhouette width given the decrease in sequence lengths from 200 to 20. As expected, the increase in the ratio of pattern length to the sequence length leads to higher ASW values. This observation implies the amount of noise in the sequences where the pattern length is relatively small. Notwithstanding the changes in ASW, the model is able to correctly cluster all sequences under different sequence lengths given external indices.

\section{\textcolor{black}{Information on Coronavirus Dataset}} \label{sec:corona_apx}
\textcolor{black}{Table~\ref{table:coronovirus} shows the detail information on the coronavirus genomes of 5 different groups together with 4 out-group genomes. The genomes are downloadable through their accession number on National Center for Biotechnology Information
Search Website\footnote{\href{https://www.ncbi.nlm.nih.gov/}{https://www.ncbi.nlm.nih.gov/}}.}

\begin{table}[!ht]
\centering
\caption{\textcolor{black}{Information on the 30 coronavirus genomes and the four non-coronavirus genomes}}\label{table:coronovirus}
\resizebox{\textwidth}{!}{
    \begin{tabular}{llclr}
    \toprule
    \textbf{Accession Number} & \textbf{Abbreviation} & \textbf{Group Number} & \textbf{Description} & \textbf{Length (bp)} \\
    \midrule
    AF304460 & 1:HCoV-229E & 1 & Human coronavirus 229E & 27317 \\
    AF353511 & 1:PEDV & 1 & Porcine epidemic diarrhea virus strain & 28033 \\
    NC\_005831 & 1:HCoV-NL63 & 1 & Human coronavirus NL63 & 27553 \\
    \hline
    AY391777 & 2:HCoV-OC43 & 2 & Human coronavirus OC43 & 30738 \\
    U00735 & 2:BCoVM & 2 & Bovine coronavirus strain Mebus & 31032 \\
    AF391542 & 2:BCoVL & 2 & Bovine coronavirus isolate BCoV-LUN & 31028 \\
    AF220295 & 2:BCoVQ & 2 & Bovine coronavirus strain Quebec & 31100 \\
    NC\_003045 & 2:BCoV & 2 & Bovine coronavirus & 31028 \\
    AF208067 & 2:MHVM & 2 & Murine hepatitis virus strain ML-10 & 31233 \\
    AF201929 & 2:MHV2 & 2 & Murine hepatitis virus strain 2 & 31276 \\
    AF208066 & 2:MHVP & 2 & Murine hepatitis virus strain Penn 97-1 & 31112 \\
    NC\_001846 & 2:MHV & 2 & Murine hepatitis virus & 31357 \\
    \hline
    NC\_001451 & 3:IBV & 3 & Avian infectious bronchitis virus & 27608 \\
    EU095850 & 3:TCoV & 3 & Turkey coronavirus isolate MG10 & 27657 \\
    \hline
    AY278488 & 4:BJ01 & 4 & SARS coronavirus BJ01 & 29725 \\
    AY278491 & 4:HKU-39849 & 4 & SARS coronavirus HKU-39849 & 29727 \\
    AY278554 & 4:CUHK-W1 & 4 & SARS coronavirus CUHK-W1 & 29736 \\
    AY282752 & 4:CUHK-Su10 & 4 & SARS coronavirus CUHK-Su10 & 29736 \\
    AY283794 & 4:SIN2500 & 4 & SARS coronavirus isolate SIN2500 & 29711 \\
    AY283795 & 4:SIN2677 & 4 & SARS coronavirus isolate SIN2677 & 29705 \\
    AY283796 & 4:SIN2679 & 4 & SARS coronavirus isolate SIN2679 & 29711 \\
    AY283797 & 4:SIN2748 & 4 & SARS coronavirus isolate SIN2748 & 29706 \\
    AY283798 & 4:SIN2774 & 4 & SARS coronavirus isolate SIN2774 & 29711 \\
    AY291451 & 4:TW1 & 4 & SARS coronavirus TW1 & 29729 \\
    NC\_004718 & 4:TOR2 & 4 & SARS coronavirus TOR2 & 29751 \\
    AY297028 & 4:ZJ01 & 4 & SARS coronavirus ZJ01 & 29715 \\
    AY572034 & 4:Civet007 & 4 & SARS coronavirus civet007 & 29540 \\
    AY572035 & 4:Civet010 & 4 & SARS coronavirus civet010 & 29518 \\
    \hline
    NC\_006577 & 5:HCoV-HKU1 & 5 & Human coronavirus HKU1 & 29926 \\
    \hline
    NC\_001564 & out:CellF & out & Cell fusing agent virus, Flaviviridae & 10695 \\
    NC\_004102 & out:HepaCF & out & Hepatitis C virus & 9646 \\
    NC\_001512 & out:NyongT & out & O’nyong-nyong virus & 11835 \\
    NC\_001544 & out:RossT & out & Ross River virus & 11657 \\
    \bottomrule
    \end{tabular}
}
\end{table}

\begin{figure}[!ht]
    \centering
    \includegraphics[width=0.8\linewidth]{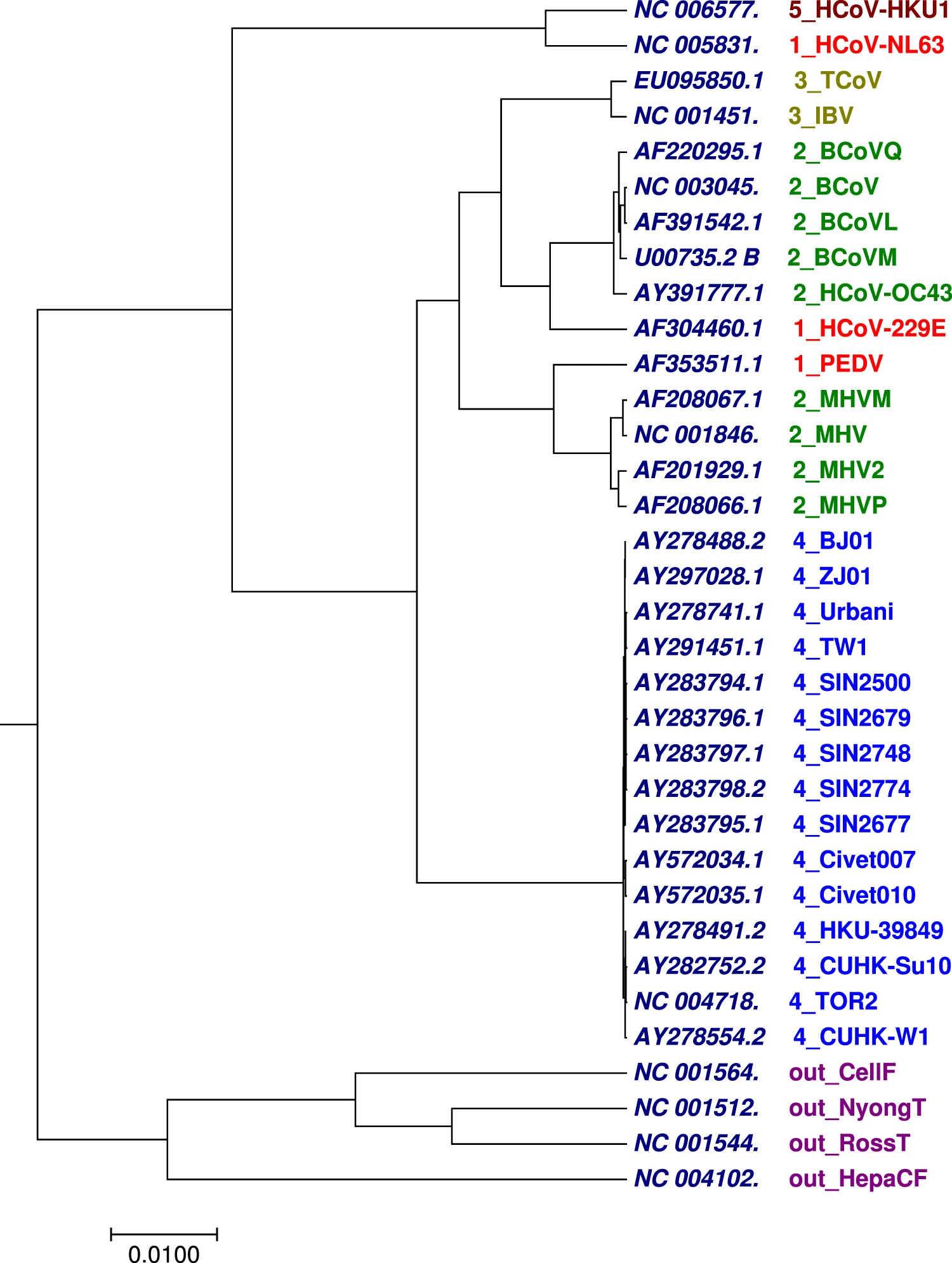}
    \caption{\textcolor{black}{The suggested phylogenetic tree of 30 coronavirus whole genomes constructed using Fuzzy integral similarity~\citep{saw2019}.}}
    \label{fig:coronavirus_old_paper}
\end{figure}

\textcolor{black}{Figure~\ref{fig:coronavirus_old_paper} shows the clustering result obtained by~\citet{saw2019}. The showed that their method outperforms $k$-mer and has advantages over feature frequency profile.}

\end{document}